\documentclass[10pt,twocolumn,letterpaper]{article}

\usepackage[pagenumbers]{cvpr} 

\usepackage{amsmath,amsfonts,bm}

\def\eqref#1{equation~\ref{#1}}

\def\1{\bm{1}}

\DeclareMathAlphabet{\mathsfit}{\encodingdefault}{\sfdefault}{m}{sl}
\SetMathAlphabet{\mathsfit}{bold}{\encodingdefault}{\sfdefault}{bx}{n}

\DeclareMathOperator*{\argmax}{arg\,max}

\usepackage{graphicx}
\usepackage{amsmath}
\usepackage{amssymb}
\usepackage{booktabs}

\usepackage{enumitem}
\usepackage[dvipsnames]{xcolor}         
\usepackage{algorithm}
\usepackage{listings}
\usepackage{float}
\usepackage{pifont}
\usepackage{subcaption}
\usepackage{multirow}
\usepackage{array,makecell,rotating}
\usepackage{wrapfig}
\usepackage{colortbl}
\usepackage{xspace}
\usepackage[accsupp]{axessibility}  

\definecolor{codegreen}{rgb}{0,0.6,0}
\definecolor{codegray}{rgb}{0.5,0.5,0.5}
\definecolor{codepurple}{rgb}{0.58,0,0.82}
\definecolor{backcolour}{rgb}{1,1,1}

\lstdefinestyle{mystyle}{
    backgroundcolor=\color{backcolour},   
    commentstyle=\color{codegreen},
    keywordstyle=\color{magenta},
    stringstyle=\color{codepurple},
    basicstyle=\ttfamily\small,
    breakatwhitespace=false,         
    breaklines=true,                 
    captionpos=b,                    
    keepspaces=true,      
    numberstyle=\tiny\color{codegray},
    numbers=left,                    
    numbersep=5pt,                  
    showspaces=false,                
    showstringspaces=false,
    showtabs=false,
    frame=tb,
    emphstyle=\ttb\color{codegreen},
}
\definecolor{darkgreen}{HTML}{119B08}
\definecolor{red2}{RGB}{252, 54, 65}
\newcommand{\green}[1]{\textcolor{darkgreen}{#1}}

\newcommand{\gcheck}{\green{\ding{52}}}

\newcommand{\arch}[1]{\textsc{#1}}
\definecolor{darkblue}{HTML}{0036A9}

\definecolor{darkred}{rgb}{0.8, 0.0, 0.0}

\def\fullname{Text-grounded Contrastive Learning\xspace}
\def\fullsmall{text-grounded contrastive learning\xspace}
\def\fullbold{\textbf{T}ext-grounded \textbf{C}ontrastive \textbf{L}earning\xspace}
\def\method{TCL\xspace}

\newcommand{\lossmethod}{$\mathcal L_\text{\method}$}
\newcommand{\lossarea}{$\mathcal L_\text{area}$}
\newcommand{\losstv}{$\mathcal L_\text{tv}$}
\renewcommand{\paragraph}[1]{\vspace{1.25mm}\noindent\textbf{#1}}

\definecolor{citecolor2}{HTML}{0071bc}
\usepackage[
    colorlinks=true,
    breaklinks=true,
    pagebackref=true,
    citecolor=citecolor2,
]{hyperref}

\usepackage[capitalize]{cleveref}
\crefname{section}{Sec.}{Secs.}
\crefname{table}{Table}{Tables}

\def\cvprPaperID{12355} 
\def\confName{CVPR}
\def\confYear{2023}

\begin{document}

\title{Learning to Generate Text-grounded Mask for \\ Open-world Semantic Segmentation from Only Image-Text Pairs}

\author{Junbum Cha
\and
Jonghwan Mun
\and
Byungseok Roh
\vspace{0.1cm}
\and
Kakao Brain\\
{\tt\small \{junbum.cha, jason.mun, peter.roh\}@kakaobrain.com}
}
\maketitle

\begin{abstract}
We tackle open-world semantic segmentation, which aims at learning to segment arbitrary visual concepts in images, by using only image-text pairs without dense annotations.
Existing open-world segmentation methods have shown impressive advances by employing contrastive learning (CL) to learn diverse visual concepts and transferring the learned image-level understanding to the segmentation task.
However, these CL-based methods suffer from a train-test discrepancy, since it only considers image-text alignment during training, whereas segmentation requires region-text alignment during testing.
In this paper, we proposed a novel \fullbold (TCL) framework that enables a model to directly learn region-text alignment.
Our method generates a segmentation mask for a given text, extracts text-grounded image embedding from the masked region, and aligns it with text embedding via \method.
By learning region-text alignment directly, our framework encourages a model to directly improve the quality of generated segmentation masks.
In addition, for a rigorous and fair comparison, we present a unified evaluation protocol with widely used 8 semantic segmentation datasets.
\method achieves state-of-the-art zero-shot segmentation performances with large margins in all datasets.
Code is available at \url{https://github.com/kakaobrain/tcl}.

\end{abstract}

\section{Introduction}

\begin{figure}[t]
    \centering
    \includegraphics[width=\columnwidth]{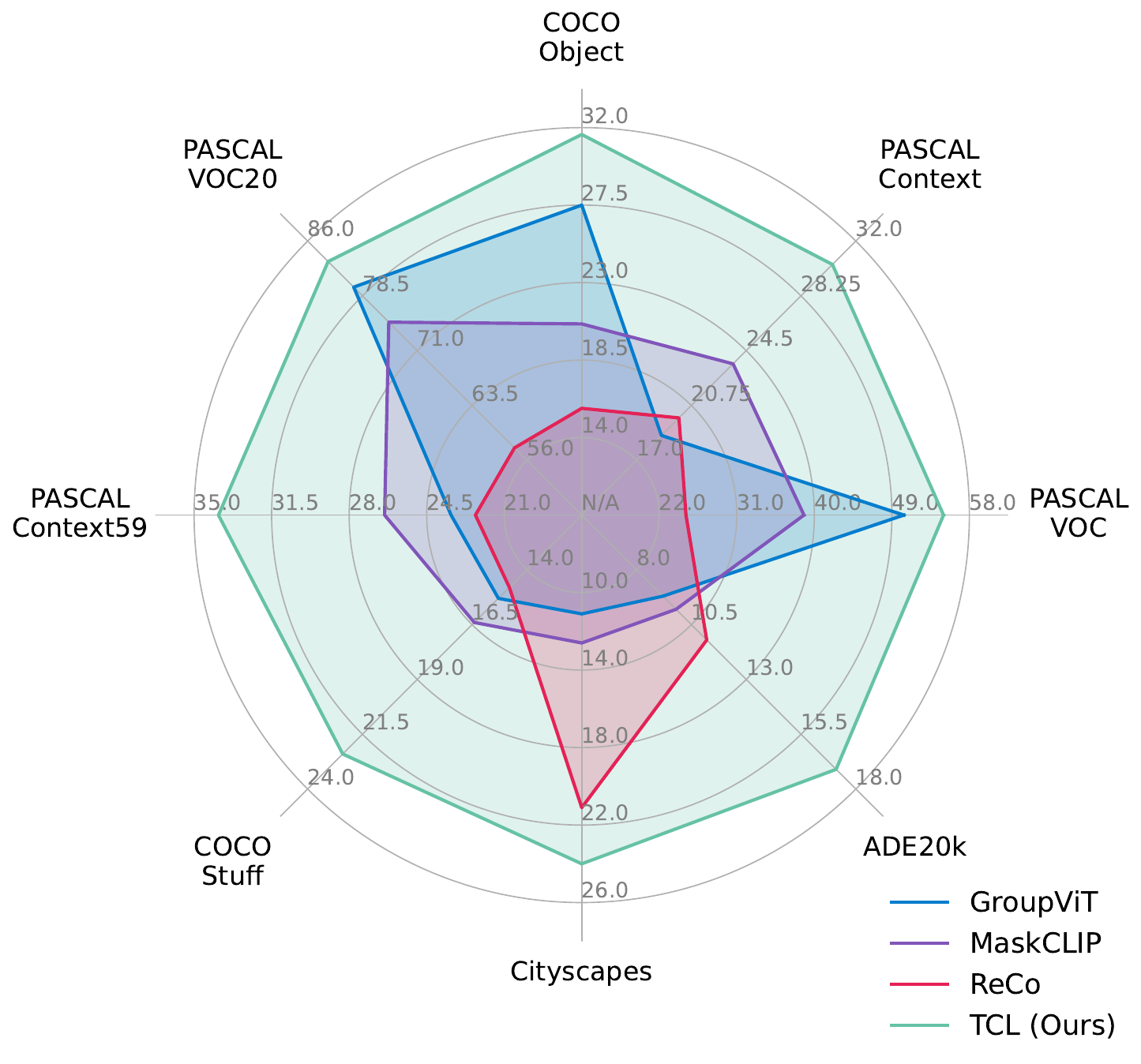}
    \vspace{-0.5cm}
    \caption{\small \textbf{Open-world segmentation performance comparison.} The proposed method remarkably outperforms existing methods in all 8 segmentation benchmark datasets.}
    \vspace{-0.2cm}
    \label{fig:radar}
\end{figure}

Open-world semantic segmentation aims to identify the arbitrary semantic concepts in the open world\footnote{This setting is often called both \textit{open-world} and \textit{open-vocabulary}. In this paper, we mainly refer to this setting as \textit{open-world} for clarity.}. 
Conventional semantic segmentation aims to learn segmentation capability for the small number of pre-defined target categories, whereas open-world semantic segmentation addresses unrestricted arbitrary categories or free-form texts.
Such segmentation capability over unlimited targets drastically extends the application scope of the open-world segmentation models. 

The first challenge for open-world segmentation is how to learn arbitrary concepts, beyond pre-defined categories.
Inspired by the success of CLIP~\cite{radford2021clip}, previous approaches~\cite{xu2022groupvit,zhou2022maskclip,shin2022reco,liu2022vilseg,ghiasi2022openseg,liang2022maskadaptedclip,li2021lseg} tackle this challenge by exploiting massive web-crawled image-text paired data;
since the texts in web-crawled data contain a global semantic description for the paired images, the large-scale image-text pairs can provide rich knowledge for arbitrary semantic categories.
However, there still remains another challenge in \emph{how to achieve precise localization of arbitrary concepts without dense annotations}.
There are several approaches that simply address this issue using dense annotation (segmentation masks) in addition to image-text pairs \cite{ghiasi2022openseg,liang2022maskadaptedclip,li2021lseg}. The dense annotation helps to improve segmentation performance in a fixed benchmark dataset, but the requirements of expensive dense annotation still limit the applicable domains and scalability of the method.

\begin{figure}[t]
    \centering
    
    \includegraphics[width=\columnwidth]{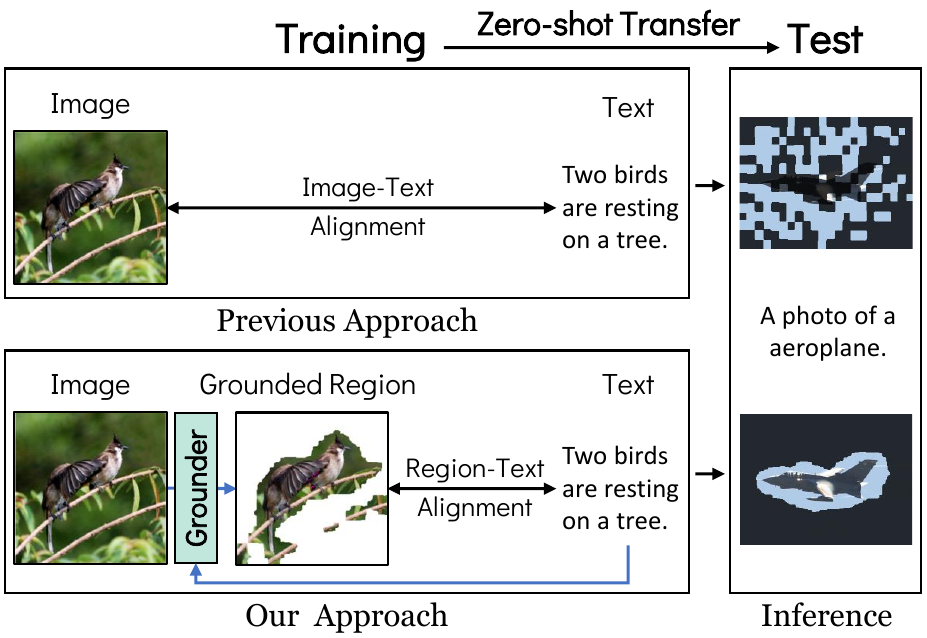}
    \vspace{-0.4cm}
    \caption{\small \textbf{A conceptual comparison between the previous approach and ours.} 
    Open-world segmentation is typically achieved through region-text alignment, which involves matching region features and text embeddings.
    However, previous methods learn image-text alignment during training, thus suffering from the alignment-level discrepancy between training and testing.
    In contrast, our method facilitates end-to-end learning of region-text alignment with only image-text pairs.
    }
    \label{fig:concept-comp}
    \vspace{-0.4cm}
\end{figure}

In this paper, therefore, we focus on open-world semantic segmentation from only image-text pairs without any dense annotation.
For this setting, the existing methods \cite{zhou2022maskclip, shin2022reco, xu2022groupvit, liu2022vilseg} learn an image-text alignment capability during training and heavily rely on the transferability of the image-text alignment to perform region-text alignment at inference.
More specifically, MaskCLIP~\cite{zhou2022maskclip} leverages CLIP models pre-trained to learn image-text alignment.
To perform region-text alignment using CLIP, MaskCLIP applies a simple heuristic modification to the CLIP image encoder.
%
GroupViT~\cite{xu2022groupvit} and ViL-Seg~\cite{liu2022vilseg} propose to cluster region-level visual features into distinct groups and generate segmentation masks by matching the groups and texts.
Note that they match the text embeddings and clustered region features in test time, but in training time, the text embeddings are aligned with global image embeddings.
%
While the existing methods have shown impressive results even through the training with image-text alignment, they still suffer from the alignment-level discrepancy between training and testing phases as depicted in \cref{fig:concept-comp}.

To address this train-test discrepancy, we propose the \fullbold (\method) framework, which allows a model to learn region-text alignment directly from the image-text pairs without any dense annotations.
Our key idea is to incorporate a text grounding procedure within contrastive learning as illustrated in \cref{fig:concept-comp},
where \method generates a segmentation mask indicating text-grounded regions, computes grounded region embeddings using the mask, and applies contrastive learning between text and grounded region.
By re-formulating the contrastive loss to be directly affected by the segmentation quality, \method enables end-to-end training of the grounder and directly improves the quality of region-text level alignment.
We also present a unified evaluation protocol using widely used 8 semantic segmentation datasets and compare existing methods in the same setting.
As a result, \method achieves state-of-the-art zero-shot segmentation performance with large margins in all datasets, as shown in \cref{fig:radar}.

Our main contributions are summarized as follows:
\begin{itemize}[leftmargin=.7cm,noitemsep,nosep]
    \item 
        We introduce a novel framework for open-world segmentation, named \fullname (\method), which enables learning region-text alignment directly without train-test discrepancy, thus learning to generate more precise segmentation masks through only image-text pairs.
    \item
        We present a unified evaluation protocol and re-evaluate recent open-world segmentation models for a fair and direct comparison.
    \item 
        We achieve the new state-of-the-art zero-shot segmentation performance on 8 segmentation datasets with large margins compared to existing methods.
\end{itemize}

\section{Related Works}
\label{sec:rel-work}

\subsection{Open-world Semantic Segmentation}

\textbf{Open-world} scenario aims to recognize arbitrary concepts in the open world. It is also called \textbf{open-vocabulary} because the target vocabulary is open rather than closed.
Contrastive Language-Image Pre-training (CLIP) \cite{radford2021clip} ushered in the era of open-world image recognition using large-scale image-text pairs \cite{kakaobrain2022coyo-700m,schuhmann2021laion,sharma2018cc3m, changpinyo2021conceptual}. 
CLIP learns the alignment between an image and a text in training time, then transfer it to the zero-shot classification by aligning image and texts indicating target classes at inference time.
The advent of CLIP enables open-world settings in various fields such as object detection \cite{gu2022vild, zhong2021regionclip}, image captioning \cite{hessel2021clipscore}, or semantic segmentation \cite{xu2022groupvit,zhou2022maskclip}.

\textbf{Open-world semantic segmentation with image-text pairs} is addressed in two different settings. 
The first is a \textbf{semi-supervised setting}, which uses dense annotation (\ie, segmentation masks) in addition to image-text pairs \cite{li2021lseg,ghiasi2022openseg,liang2022maskadaptedclip}. Semi-supervised approaches learn segmentation capability using dense annotation and expand the target vocabulary using image-text supervision.
LSeg \cite{li2021lseg} expands target class vocabulary using image-label datasets and CLIP text encoder \cite{radford2021clip}. OpenSeg \cite{ghiasi2022openseg} and OVSeg \cite{liang2022maskadaptedclip} first train a mask generator using dense annotation and expand target vocabulary using image-text datasets. 
The use of dense annotation makes the model learn
region-level alignment instead of image-level alignment, leading to high-quality segmentation masks.
However, it still relies on costly dense annotation, and applicable domains are limited to the domains where dense annotation is available.

The target of this paper is an \textbf{unsupervised setting}, which aims to learn segmentation from only image-text pairs without any dense annotation \cite{zhou2022maskclip,shin2022reco,xu2022groupvit,liu2022vilseg}.
Since the massive image-text pairs are easily obtained by web crawling without human annotators, applicable domains of unsupervised methods become almost unlimited. 
In order to achieve segmentation capability using only image-text pairs, we need to learn region-text alignment instead of image-text alignment and train a text-grounded mask generator.
However, the absence of dense annotation makes this approach challenging.
Existing open-world semantic segmentation studies have taken a strategy to bypass this issue. 
Instead of learning region-level alignment directly, they transfer image-level alignment to region-level by heuristic modification \cite{zhou2022maskclip,shin2022reco} or clustering \cite{xu2022groupvit,liu2022vilseg}.
MaskCLIP \cite{zhou2022maskclip} proposes to obtain a dense image embedding from CLIP image encoder through heuristic modification of the last attention layer. 
Even though it has several limitations, such as low output resolution or noisy segmentation results, they show it is a simple yet effective way to obtain an initial segmentation map for refinement. 
ReCo \cite{shin2022reco} proposes an advanced refinement method based on MaskCLIP, by retrieval and co-segmentation.
Clustering-based methods \cite{xu2022groupvit,liu2022vilseg} learn representations using CL with image-text pairs.
They compute region-level image embedding by clustering sub-region embeddings.
These approaches also have shown impressive results but have several limitations: ($i$) the learning objective is still image-level alignment due to lack of the region annotation, ($ii$) the number of clusters is pre-defined independent of the given image, and ($iii$) clustering sub-region image embeddings is independent of the query text. 
In summary, existing methods indirectly address region-level alignment problems by learning image-level alignment.
To tackle this problem, we propose a novel region-level alignment objective, named \fullname (\method).

\subsection{Region-level Contrastive Learning}

Learning region-level alignment instead of image-level alignment is a fundamental target objective in dense tasks, such as segmentation or object detection. There are approaches to learn region-level alignment using dense annotation in the semi-supervised setting. They first train mask or region proposal networks using dense annotation and learn alignment between the proposals and texts \cite{ghiasi2022openseg,zhong2021regionclip,kamath2021mdetr}.
For example, OpenSeg \cite{ghiasi2022openseg} trains a class-agnostic mask generator using dense annotation.
In the object detection field, RegionCLIP \cite{zhong2021regionclip} employs an off-the-shelf region proposal network and learns region-level alignment.
In contrast to the existing region-level methods, the proposed method learns region-level alignment without any dense annotation.

\section{Methods}
\label{sec:methods}


\begin{figure*}[t]
    \centering
    \includegraphics[width=0.95\linewidth]{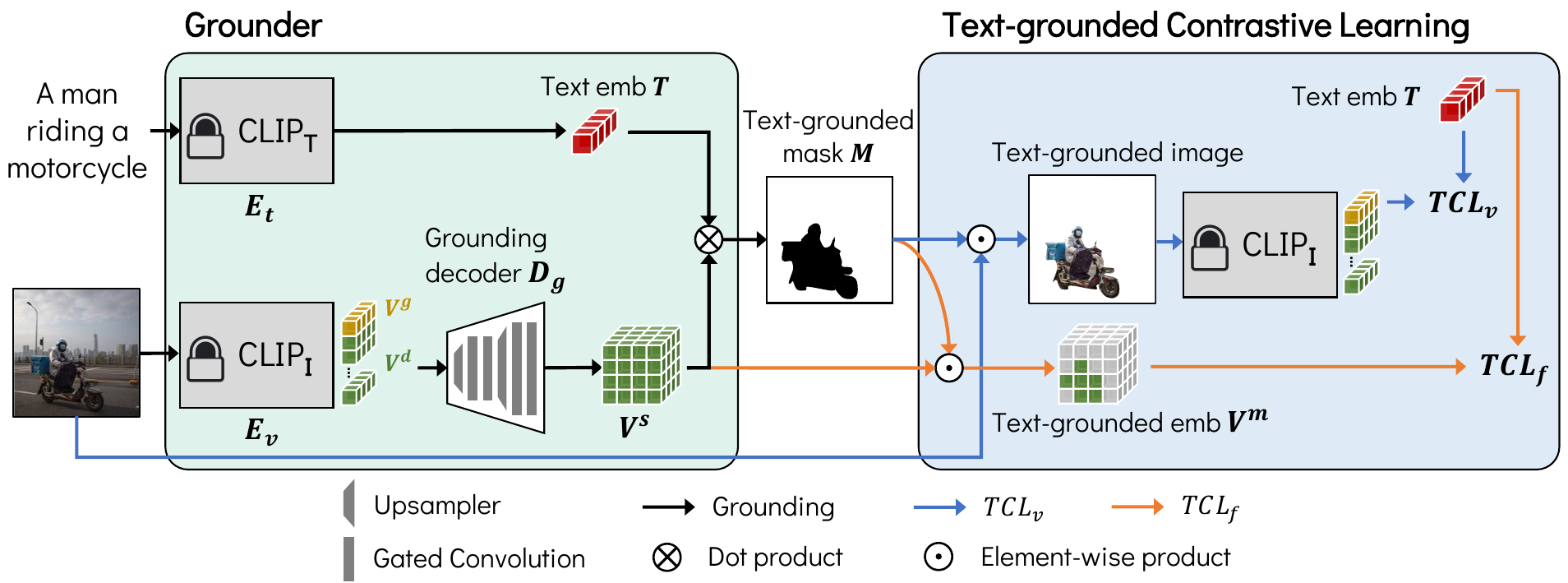}
    \caption{\small \textbf{Overall training pipeline of \method.} 
    The proposed \method framework first obtains text-grounded masks using a grounder and then learns the grounder using \fullsmall. 
    By incorporating the text grounding process with contrastive learning, our framework can directly learn region-text alignment that is required for precise segmentation.
    \texttt{CLIP\textsubscript{T}} and \texttt{CLIP\textsubscript{I}} indicate CLIP text and image encoders, respectively. CLIP encoders are frozen and we train the grounding decoder only. After training, the \method block is discarded, and only the Grounder block is used to generate the text-grounded segmentation mask for inference.}
    \label{fig:overallpipeline}
    \vspace{-0.3cm}
\end{figure*}


\subsection{Overview}

Open-world semantic segmentation is a task that aims to learn a model capable of zero-shot segmentation for arbitrary visual concepts, not restricted to pre-defined ones.
Our main goal is to develop an open-world segmentation algorithm using only image-text pairs.
However, achieving this objective is challenging because there is no explicit supervision (\ie, pixel-level dense annotations) for text-described region segmentation.
Existing methods learn models parametrized by $\theta$ to maximize the mutual information between paired images and texts~\cite{oord2018infonce,radford2021clip} as follows:
\begin{equation}
    \argmax_{\theta} I_{\theta}(\mathbf x^V; \mathbf x^T),
\end{equation}
where $( \mathbf{x}^V, \mathbf{x}^T )$ is a random image and text pair.
This objective encourages the model to learn the alignment between images and texts, 
however, at test time, the learned model generates segmentation masks for arbitrary concepts by computing region-text alignments.
Such alignment-level discrepancy between train and test time can lead the model to a sub-optimal solution as shown in \cref{fig:concept-comp}.
With this in consideration, to bridge the gap between the objective of conventional contrastive learning (CL) and the requirement of the zero-shot segmentation, we propose \fullname (\method) which incorporates a text grounding process within CL to enable learning region-text alignment directly.
As a text grounding module, we introduce a grounder to generate segmentation masks for the given texts.
In a nutshell, \method learns a model to maximize mutual information between text-grounded regions and texts as follows:
\begin{equation}
    \argmax_{\theta} I_{\theta}(\mathbf m \cdot \mathbf x^V; \mathbf x^T),
\end{equation}
where $\mathbf m$ is a text-grounded mask of random variable indicating the text-described region.
Compared to contrastive learning that implicitly learns a grounding capability, \method has a clear advantage of explicitly learning the grounding capability through the end-to-end trainable grounder.

In the rest of this section, we first explain the text-grounded mask generation procedure by the grounder.
Then, we describe how we define losses using the generated mask to train our open-world grounder with text-grounded contrastive learning.
Lastly, we explain how our model performs zero-shot inference for arbitrary concepts.

\subsection{Grounder}

\cref{fig:overallpipeline} illustrates our overall training pipeline. 
For an input batch of paired texts $\mathbf X^T$ and images $\mathbf X^V$, \method first performs a grounding process to identify text-grounded regions for a text via a grounder. 
The grounder consists of three components: ($i$) image encoder $E_v$ is in charge of providing a single (L2-normalized) global feature as well as dense patch-level features, ($ii$) text encoder $E_t$ provides a (L2-normalized) text embedding feature, and ($iii$) grounding decoder $D_g$ converts dense features from image encoder into finer pixel-level embeddings for alignment with text.
In practice, we adopt a pre-trained CLIP model \cite{radford2021clip} to initialize two encoders and freeze them to preserve and exploit the rich knowledge of CLIP learned during large-scale pre-training.
The text-grounded masks are computed by the position-wise dot product between text embedding and dense pixel-level embedding. The overall process of grounder is summarized as follows:
\begin{align}
    &\mathbf T = E_t\left(\mathbf X^T \right)\quad\text{and}\quad
    \mathbf V^g, \mathbf V^d = E_v \left( \mathbf X^V \right), \\
    &\mathbf V^s = D_g(\mathbf V^d), \\
    &\mathbf M_{i,j} = \sigma\left(w\cdot \mathbf t_j^\top \mathbf V^s_i+b\right), \label{eq:pred-mask}
\end{align}
where $\mathbf T \in \mathbb R^{B \times C}$, $\mathbf V^g \in \mathbb R^{B\times C}$, and $\mathbf V^d \in \mathbb R^{B \times L \times C}$ are normalized text embeddings, normalized global image embeddings, and dense image features from CLIP encoders, $\mathbf V^s \in \mathbb R^{B\times C\times H\times W}$ is normalized pixel-level dense embeddings by the grounding decoder, and $\mathbf M \in \mathbb R^{B\times B\times H\times W}$ is text-grounded masks between images and texts in the batch.
$B$, $C$, and $L$ indicate a batch size, the embedding dimension size, and the number of patches, respectively.
$\sigma$ is a sigmoid function, and $w, b$ are learnable scalar projection. 

The generated text-grounded masks are used to extract text-grounded image embedding.
By replacing the global image embedding with text-grounded image embedding in the contrastive learning framework, \method enables the model to learn region-text alignment in an end-to-end manner.
In the following section, we describe how the generated mask $\mathbf M$ is used for text-grounded contrastive learning.


\subsection{\fullname}
\label{subsec:training-objectives}

Recall that the main idea of TCL is to use text-grounded images instead of whole images, unlike conventional CL.
For this purpose, we define \method losses in three different levels---image-level, feature-level, and area-level---using the generated masks $\mathbf{M}$ for all pairs of images and texts in a batch; the detailed pseudo code to compute \method losses is given in Appendix~\ref{sec:pseudo-code}.
We also employ smooth regularization to further improve the quality of generated masks.


\paragraph{Image-level \method loss.}
One intuitive way to compute the text-grounded image embedding is to encode only the regions of the image that contains the semantics of the paired text, using the image encoder.
%
To make the whole process end-to-end trainable, we compute a differentiable masked image by multiplying the given image $\mathbf{X}^V_i$ and a binarized mask $\mathbf{M}^b_{i,i}$ obtained from the generated mask $\mathbf{M}_{i,i}$ using Gumbel-Max~\cite{jang2017gumbelsoftmax}.
The masked image is then fed into the image encoder, $\tilde{\mathbf{v}}^g_i, \tilde{\mathbf{v}}^d_i = E_v\left( \mathbf M^b_{i,i}\cdot \mathbf X_i^V \right)$, to obtain text-grounded image embedding $\tilde{\mathbf{v}}^g_i$.
We compute the cosine similarity matrix between text-grounded image embeddings and text embeddings in a batch by $S^m_{i,j} = {{}\tilde{\mathbf v}^g_i}^\top \mathbf t_j$.
Finally, we use the symmetric version of InfoNCE~\cite{oord2018infonce,radford2021clip} to define the image-level TCL loss $\mathcal{L}_{\text{\method}_v}$ to make the representations of positive image-text pairs similar to each other while the representations of negative pairs dissimilar based on the similarity matrix:
%
%
\begin{equation}
    \mathcal L_{\text{\method}_v} = \text{InfoNCE}\left( \mathbf S^m \right),
\end{equation}
\begin{align}
    \text{InfoNCE} \left( \mathbf S \right) = &-\frac1{2B} \sum^B_i \log \frac{\exp({S}_{i,i}/\tau)}{\sum^B_j \exp({S}_{i,j}/\tau)} \nonumber \\ 
    &- \frac1{2B} \sum^B_i \log \frac{\exp({S}_{i,i}/\tau)}{\sum^B_j \exp({S}_{j,i}/\tau)},
\end{align}
where $\tau$ is a learnable temperature.

\paragraph{Feature-level \method loss.}
The image-level \method loss drives a model to generate segmentation masks for the paired texts (\ie, texts of positive pairs). However, we observe that this loss alone is insufficient to prevent the model from generating masks for regions not described in the text, particularly for salient regions.
This raises the need to suppress negative masks obtained from unrelated texts (\ie, texts of negative pairs), but computing the image-level TCL loss for negative masks is infeasible due to the high computational cost of encoding text-grounded images.
To overcome this challenge, we introduce the feature-level \method loss, which enables the effective computation of features of the negative masks.
Specifically, for pixel-level dense embeddings $\mathbf V^s_i \in \mathbb R^{C\times H\times W}$ from grounding decoder and a text embedding vector $\mathbf t_j$, we compute feature-level text-grounded image embedding $\mathbf v^f_{i,j} \in \mathbb R^{C}$ by:
\begin{equation}
    \mathbf v^f_{i,j} = \frac{\sum_{h,w} {M}_{i,j,h,w} \cdot \mathbf v^s_{i,:,h,w}}{\sum_{h,w} {M}_{i,j,h,w}}.
\end{equation}
%
Note that this feature-level embedding is computed using negative masks $\mathbf M_{i,j ~(i \neq j)}$, different from the image-level TCL loss.
We then compute the cosine similarity $S^f_{i,j}={\mathbf v^f_{i,j}}^\top \mathbf t_j$ between all pairs of text embeddings and feature-level text-grounded image embeddings in the batch. The feature-level TCL loss is defined as follows:
%
\begin{equation}
    \mathcal L_{\text{\method}_f} = \text{InfoNCE}\left( \mathbf S^f \right).
\end{equation}



\paragraph{Area TCL loss.}
The image-level and feature-level TCL losses focus on generating a mask to capture the text-described region in the image.
However, the model can collapse into a trivial solution with only these losses---generating a mask for the entire image instead of the desired region. To prevent this collapse, we introduce an additional objective to our TCL framework, named area TCL loss, which incorporates priors on the mask area to ensure capturing only the text-described region.
To be specific, for the positive masks (masks from positive pairs) $\mathbf M^+$ and the negative masks (masks from negative pair) $\mathbf M^-$, we denote the area of positive and negative masks by $\overline{\mathbf M^+}$ and $\overline{\mathbf M^-}$, respectively. 
The area TCL loss is defined by L1-distance between the area priors and the expected area of each mask:
\begin{equation}
    \mathcal L_{\text{area}} = \left\|p^+ - \mathbb E \left[ \overline{\mathbf M^+} \right] \right\|_1 + \left\| p^- - \mathbb E \left[ \overline{\mathbf M^-} \right] \right\|_1,
\end{equation}
where $p^+$ and $p^-$ are positive and negative area priors. 
For the negative area prior $p^-$, intuitively, we can expect the area of the negative masks to be 0.0. 
We set the positive area prior $p^+$ to 0.4, which is the average text-described region area measured by MaskCLIP \cite{zhou2022maskclip} in the CC3M dataset \cite{sharma2018cc3m}.

\paragraph{Smooth regularization.}
In the image-text dataset, a text usually describes the salient object or concept in the paired image. We observe that the regions described by the text are generally smooth rather than noisy.
We employ total variation (TV) regularization loss \cite{rudin1992totalvariation} to incorporate this smoothness observation in the objective.
The TV loss is applied to both mask and pixel-level dense embedding:
\begin{equation}
    \mathcal L_\text{tv}=\| \mathbf M \|_\text{TV} + \|\mathbf V^s \|_\text{TV},
\end{equation}
where $\| \cdot \|_\text{TV}$ is the anisotropic TV norm.

\paragraph{Final loss.}
Our final loss function is defined by:
\begin{equation}
    \mathcal L=\underbrace{\lambda_\text{\method} \mathcal L_\text{\method} + \lambda_{\text{area}} \mathcal L_{\text{area}}}_{\text{TCL losses}} + \underbrace{\lambda_\text{tv} \mathcal L_\text{tv}}_{\text{regularization}},
\end{equation}
where $\mathcal L_\text{\method} = \mathcal L_{\text{\method}_v} + \mathcal L_{\text{\method}_f}$, and $\lambda_{\text{\method}}$, $\lambda_{\text{area}}$, $\lambda_{\text{tv}}$ are hyperparameters to balance three losses. 

\subsection{Inference Pipeline}
The zero-shot inference pipeline is similar to CLIP \cite{radford2021clip}, except for performing pixel-level classification instead of image-level classification. 
Specifically, for text embeddings $\mathbf T \in \mathbb R^{N \times C}$ and a pixel-level dense embedding $\mathbf v^s \in \mathbb R^{C \times H \times W}$, text-grounded mask $\mathbf M \in \mathbb R^{N \times H \times W}$ is computed by \cref{eq:pred-mask}, where $N$ is the number of target classes. The final segmentation map $\mathcal M$ is computed by:
\begin{equation}
    \mathcal M_{h,w} = \argmax_n M_{n,h,w}.
\end{equation}
Prompt templates such as ``\texttt{a photo of a \{label\}.}'' are used to generate text embeddings as in CLIP \cite{radford2021clip}.



\section{Experiments}
\label{sec:experiments}

\subsection{Experiment Settings}
\label{subsec:experiment-settings}
\begin{table*}[t]
\centering
\small
\renewcommand{\arraystretch}{1.0}
\setlength{\tabcolsep}{4pt}
\begin{tabular}{l|ccc|ccccc|c}
\toprule
                            & \multicolumn{3}{c|}{{with background class}}        & \multicolumn{5}{c|}{{without background class}} &                                                 \\
\textbf{Methods}             & {VOC}  & {Context} & {Object}    & {VOC20} & {Context59} & {Stuff} & {City} & {ADE} & Avg.  \\
\midrule
GroupViT \scriptsize{(YFCC)}    & 49.5             & 19.0             & 24.3                 & 74.1           & 20.8               & 12.6                & 6.9                 & 8.7              & 27.0\\
GroupViT \scriptsize{(RedCaps)} & \underline{50.4} & 18.7             & \underline{27.5} & \underline{79.7}           & 23.4               & 15.3                & 11.1                & 9.2      & \underline{29.4}              \\
MaskCLIP$^\dagger$                &  29.3            & 21.1             & 15.5             & 53.7           & 23.3               & 14.7                & \underline{21.6}                & 10.8     & 23.7        \\
MaskCLIP                        &  38.8            & \underline{23.6} & 20.6            & 74.9           & \underline{26.4}               & \underline{16.4}                & 12.6                & 9.8  & 27.9            \\
ReCo                            &  25.1            & 19.9             & 15.7            & 57.7           & 22.3               & 14.8                & 21.1                & \underline{11.2}      & 23.5     \\
\midrule
\multirow{2}{*}{\method (Ours)}               & \textbf{55.0} & \textbf{30.4}    & \textbf{31.6}        & \textbf{83.2}  & \textbf{33.9}      & \textbf{22.4}       & \textbf{24.0}       & \textbf{17.1}  & \textbf{37.2}  \\
               & \textbf{(\green{+4.6})} & \textbf{(\green{+6.8})}    & \textbf{(\green{+4.1})}        & \textbf{(\green{+3.5})}  & \textbf{(\green{+7.5})}      & \textbf{(\green{+6.0})}       & \textbf{(\green{+2.4})}       & \textbf{(\green{+5.9})}  & \textbf{(\green{+7.8})}  \\
\bottomrule
\end{tabular}
\vspace{-0.1cm}
\caption{\small \textbf{Zero-shot segmentation performance comparison on 8 semantic segmentation datasets.} mIoU metric is used in every experiment. 
We highlight \textbf{the best} and \underline{second-best} results.
MaskCLIP$^\dagger$ indicates their baseline method without additional refinement techniques. The YFCC and RedCaps of GroupViT indicate their training datasets in addition to CC12M. Each dataset abbreviation stands for VOC: PASCAL VOC, Context: PASCAL Context, Object: COCO-Object, Stuff: COCO-Stuff, City: Cityscapes, ADE: ADE20K.
}
\label{table:main}
\vspace{-0.3cm}
\end{table*}


\paragraph{Unified evaluation protocol.}
In open-world semantic segmentation, a standard evaluation protocol is not yet established.
Previous studies conduct an evaluation using their own protocols such as different data processing strategies on different datasets \cite{xu2022groupvit,shin2022reco,zhou2022maskclip,liu2022vilseg}; 
surprisingly, even for the same dataset, the target classes are sometimes different across studies.
For a fair comparison, we present a unified evaluation protocol following the open-world scenario where prior access to the target data before evaluation is not allowed.
Under this scenario, the proposed protocol prohibits dataset-specific hyperparameters or tricks, \eg, class name expansion or rephrasing, leading to performance overestimation.
For example, we observe that \method can get significant performance gains by expanding the target class of ``person'' to its sub-concepts (\eg, man, woman, worker, rider, etc.), but the kinds of class name-based tricks are not allowed in our unified evaluation protocol because the expansion depends on the target class names.
With this consideration, we evaluate models using unified class names from the default version of MMSegmentation \cite{mmseg2020} without class name-based tricks.
Dense CRF \cite{krahenbuhl2011dcrf} is not used identically due to its expensive computational cost. All other evaluation settings follow GroupViT \cite{xu2022groupvit}, where the input image is resized to have a shorter side of 448.
We employ mean intersection-over-union (mIoU) as a performance metric, which is a standard metric in semantic segmentation.
While we aim to provide a fair comparison, defining fair conditions can be subjective. Thus, we provide further results and discussion on this topic in \cref{sec:fair_comparison}, especially regarding dataset scale and refinement methods.

\paragraph{Benchmark datasets and comparison methods.}
We provide an extensive evaluation on widely used 8 benchmarks, categorized into two groups: ($i$) with background class (PASCAL VOC \cite{everingham2010pascal}, PASCAL Context \cite{mottaghi2014context59}, and COCO-Object \cite{caesar2018cocostuff}), and ($ii$) without background class (PASCAL VOC20 \cite{everingham2010pascal}, PASCAL Context59 \cite{mottaghi2014context59}, COCO-Stuff \cite{caesar2018cocostuff}, Cityscapes \cite{cordts2016cityscapes}, and ADE20K \cite{zhou2019ade20k}).
Note that open-world segmentation methods rely on the textual description of class names, which may require additional considerations for the background class, such as probability thresholding instead of using the ``background'' description as is. The datasets with background class evaluate this aspect.
We compare \method with all existing open-sourced methods, including GroupViT~\cite{xu2022groupvit}, MaskCLIP~\cite{zhou2022maskclip}, and ReCo~\cite{shin2022reco} under the unified protocol. We also include their variants in comparison baselines for an extensive comparison.
Additional details and comparisons are given in \cref{sec:additional-comparison-and-details}.

\paragraph{Implementation details.}
For the grounder, we use the CLIP ViT-B/16 model where the size of input images is $224 \times 224$ and the patch size is $16 \times 16$. 
Following MaskCLIP~\cite{zhou2022maskclip}, we modify the last attention layer of the CLIP image encoder to acquire the dense embedding representing local semantics.
The grounding decoder consists of four gated convolution blocks with two upsampling interpolations, and
we use pixel-adaptive mask refinement (PAMR) \cite{araslanov2020pamr} for mask refinement. 
%
Further details on the model architecture are provided in \cref{sec:arch-details}.
We use CC 3M and 12M datasets \cite{sharma2018cc3m,changpinyo2021conceptual} for training.
The loss weights of $\lambda_\text{\method}=0.1, \lambda_\text{area}=0.4, \lambda_\text{tv}=1.0$ are used. We train the model with a batch size of $1024$ and a learning rate of $7.5 \times 10^{-5}$ for total $50,000$ iterations with $15,000$ warmup steps and cosine schedule. AdamW optimizer \cite{loshchilov2018adamw} is used with a weight decay of $0.05$.

\subsection{Zero-shot Transfer to Semantic Segmentation}
\paragraph{Comparison of existing methods.}
We extensively compare existing open-world semantic segmentation methods in \cref{table:main} using the proposed unified protocol, including two checkpoints of GroupViT \cite{xu2022groupvit} and two variants of MaskCLIP \cite{zhou2022maskclip}. 
%
Between the existing methods, GroupViT achieves the best average performance, particularly on object-oriented datasets such as VOC, VOC20, and COCO-Object. However, its performance tends to decrease when the target dataset is dominated by stuff classes. 
%
On the other hand, MaskCLIP performs the best on stuff-oriented datasets such as Context, Context59, and COCO-Stuff, benefiting from the large-scale pre-trained CLIP model. 
We conjecture it benefits from leveraging a large-scale pre-trained CLIP model. 
The refinement techniques proposed in MaskCLIP \cite{zhou2022maskclip} improve the average performance but significantly degrade it on Cityscapes ($21.6 \rightarrow 12.6$), suggesting the limitation of the heuristic refinement methods.
%
The significant performance degradation of MaskCLIP and ReCo between VOC20 and VOC may imply the need for consideration for background class.

\paragraph{\method remarkably outperforms existing methods.}
Although the performances of existing methods vary depending on the characteristics of the evaluation datasets, \method outperforms all the other methods by large margins across all datasets as shown in \cref{table:main}.
These results demonstrate that our \method framework successfully addresses the alignment-level train-test discrepancy that exists in the previous methods by learning the region-level alignment. 
In addition, region-level alignment learning of \method allows our model to learn the capability to distinguish the background region in a data-driven manner, thus, our method can address the background class without any heuristic post-processing that the previous methods typically rely on.




\subsection{Qualitative Results}
\label{subsec:qualitative-results}

\begin{figure}[tb]
\begin{center}
\includegraphics[width=\columnwidth]{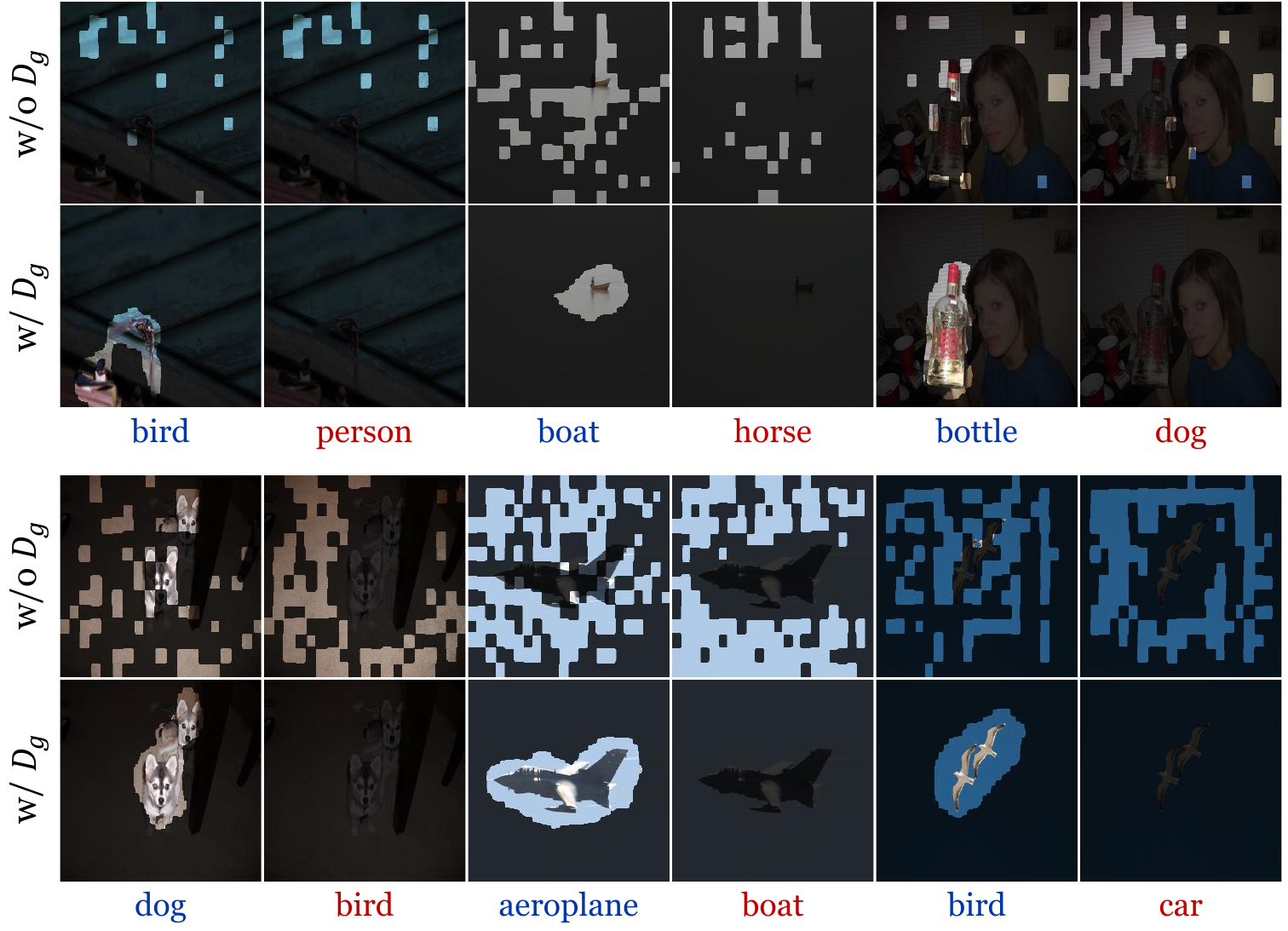}
\end{center}
\vspace{-0.4cm}
\caption{
    \small \textbf{Visualization of the generated text-grounded masks.}
    ``w/o $D_g$'' rows show the generated text-grounded masks without the grounding decoder ($D_g$), \ie, CLIP dense features $\mathbf V^d$ are used instead of pixel-level dense embeddings $\mathbf V^s$. ``w/ $D_g$'' rows show the results with the grounding decoder. 
    Each image is compared using both \textcolor{darkblue}{positive} (blue) and \textcolor{darkred}{negative} (red) prompts.
    The results show that the grounder accurately and finely captures the text-described region with less noise via the grounding decoder.
}
\label{fig:grounding_viz}
\vspace{-0.3cm}
\end{figure}
\paragraph{Visualization of the generated text-grounded masks.}
\cref{fig:grounding_viz} illustrates the impact of the learned grounding decoder.
Since we follow MaskCLIP \cite{zhou2022maskclip} modification, the results in ``w/o $D_g$'' rows can be regarded as the initial results of MaskCLIP before refinement.
Despite the vast pre-training scale and remarkable zero-shot classification performance of CLIP \cite{radford2021clip}, its grounding capability is limited because the learning objective targets image-level alignment (See ``w/o $D_g$'' rows). 
In contrast, the grounding decoder ($D_g$) learns the region-level alignment by \method, resulting in more precise, finer, and less noisy generated masks (See ``w/ $D_g$'' rows). 

\begin{figure*}[t]
    \centering
    \begin{subfigure}{0.6\linewidth}
        \includegraphics[width=\columnwidth]{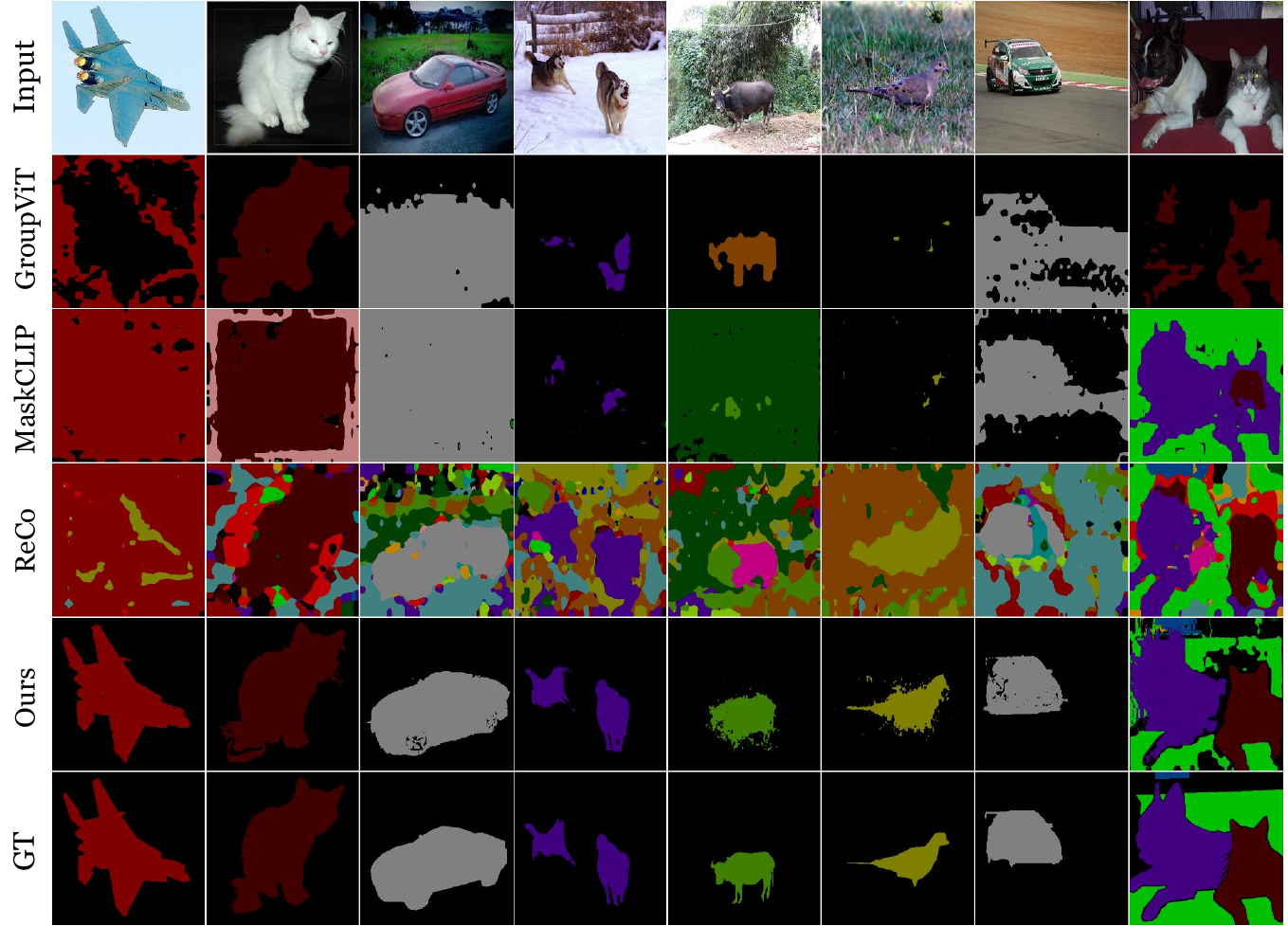}
        \caption{\small \textbf{Examples in PASCAL VOC.}}
        \label{fig:qualitative-voc}
    \end{subfigure}
    \begin{subfigure}{0.381\linewidth}
        \includegraphics[width=\columnwidth]{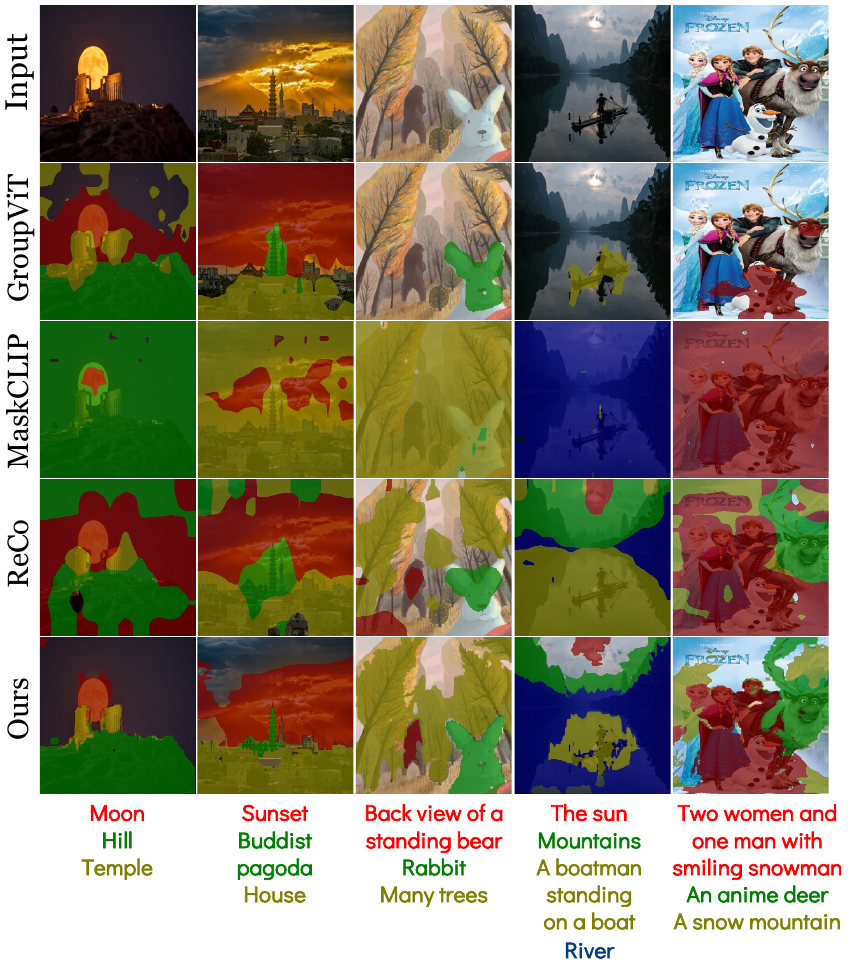}
        \caption{\small \textbf{Examples in the wild.}}
        \label{fig:qualitative-web}
    \end{subfigure}
    \vspace{-0.1cm}
    \caption{\small 
    (a) The comparison shows the error types of each method in the VOC dataset. GroupViT tends to make an error on a large group rather than noisy results. ReCo suffers from segmentation of the background region. MaskCLIP tends to fail at capturing the target area precisely.
    (b) shows results on the wild web images and free-form texts. Texts used as target classes are shown at the bottom of the images. 
    }
    \label{fig:qualitative}
    \vspace{-0.2cm}
\end{figure*}

\paragraph{Qualitative comparison.}
We qualitatively compare the proposed method in \cref{fig:qualitative}. 
On the PASCAL VOC dataset (\cref{fig:qualitative-voc}), we observe various types of errors in each comparison method. 
The grouping procedure of GroupViT \cite{xu2022groupvit} makes the segmentation results less noisy, but it also causes an incorrect segmentation of a large group. ReCo \cite{shin2022reco} struggles with the segmentation of background regions due to the lack of consideration about the background class. MaskCLIP \cite{zhou2022maskclip} does not take this into account as well, but its refinement methods make the results less noisy.
%
%
In addition, we present examples in the wild to show open-world segmentation capability in \cref{fig:qualitative-web}.
We collect test samples containing visual concepts not included in conventional segmentation datasets (\eg, moon, sunset) or free-form texts (\eg, ``two women and one man with a smiling snowman").
GroupViT tends to focus on the main object of the image and regard the other objects as background, which is consistent with its good performance in object-oriented datasets. 
Interestingly, in this qualitative comparison in the wild, we observe ReCo consistently outperforms MaskCLIP contrary to the quantitative results. 
We conjecture that this is because the refinement approach of ReCo is data-driven, while the refinement approach of MaskCLIP depends on heuristic post-processing, which may not guarantee general improvement.
Compared to the baselines, \method consistently generates more precise segmentation masks. 
These results demonstrate that our proposed method, which learns region-level alignment, improves the segmentation quality both in the evaluation dataset and in web images in the wild.
Additional qualitative results are provided in \cref{sec:additional-qualitative}.

\paragraph{Additional analysis on failure cases and model behavior} are provided in \cref{sec:failure-case-study,sec:behavior-analysis}, respectively.

\subsection{Ablation Studies}
We investigate the impact of individual components of the proposed framework by ablation studies on the training split of the PASCAL VOC20 dataset.
We use a short learning schedule with a batch size of 512 for total $40,000$ iterations including $10,000$ warmup steps.

\paragraph{Baseline to TCL.}
\cref{table:ablations-baseline-to-tcl} presents cumulative ablation studies on the grounding decoder and the TCL losses. 
Our initial model before training based on MaskCLIP~\cite{zhou2022maskclip} is referred to as the baseline (\arch{a}), which modifies the last attention layer of the CLIP image encoder.
When we add only the grounding decoder to the baseline without TCL loss (\arch{b}), there is no improvement in performance. 
This suggests that training the decoder with the same CL loss as the pre-training (CLIP) does not enhance the localization capabilities.
As shown in (\arch{c}), the proposed framework becomes complete with TCL loss.

\paragraph{Impact of individual TCL losses.}
The influence of each component of the proposed \method loss and its effect on the segmentation performance compared to the conventional CL loss are shown in \cref{table:ablations-gcl}.
Smooth regularization is used for all experiments in this table.
The CL loss (\arch{d}) is computed by applying attention pooling \cite{radford2021clip} to the dense image embedding $\mathbf V^s$.
When comparing (\arch{d}) and (\arch{c}), the proposed \method loss remarkably improves the segmentation performance ($61.1
 \rightarrow 77.4$). Image-level or feature-level \method loss (\arch{e}, \arch{f}) solely improves the performance significantly, and using both losses together provides further performance gain. 
 Using CL in addition to TCL (\arch{g}) does not improve performance, and it is essential to use area TCL loss in TCL framework to prevent model collapse (\arch{h}), as described in \cref{subsec:training-objectives}.
The difference between (\arch{b}) and (\arch{d}) is the use of smooth regularization.

\paragraph{Hyperparameters.}
\cref{table:ablations-TCL,table:ablations-area,table:ablations-tv} shows the performance changes according to the variation of the loss weight hyperparameters (HPs).
The first rows show the importance of each loss ($\lambda=0.0$ cases). 
The absence of area TCL loss causes a significant performance drop (\cref{table:ablations-area}), as mentioned above.
Smooth regularization also significantly contributes to the final performance (\cref{table:ablations-tv}), supporting our assumption that the text-described region is smooth rather than noisy.
Note that the sensitivity on HPs is about loss balancing, not about the target dataset. 
As an open-world segmentation method, \emph{TCL does not require any tuning with the target dataset, including model fine-tuning and inference HPs tuning}. Once a TCL model is trained, we evaluate the model for every benchmark without any fine-tuning.

\newcommand\scale{1.0}
\definecolor{graycolor}{gray}{.873}
\newcommand\grayrow{\rowcolor[gray]{0.873}}
\newcommand{\head}[1]{\small{#1}}
\newcommand\voc{\head{VOC20}}
\newcommand\ctx{\head{Ctx59}}
%
\newlength\savewidth\newcommand\shline{\noalign{\global\savewidth\arrayrulewidth
  \global\arrayrulewidth 1pt}\hline\noalign{\global\arrayrulewidth\savewidth}}

\begin{table}
    \centering
    \subfloat[
        {\small \textbf{Baseline to TCL}.}
        \label{table:ablations-baseline-to-tcl}
    ]{
        \begin{minipage}{0.42\columnwidth}
        \centering
        \small
        \renewcommand{\arraystretch}{1.05}
        \setlength{\tabcolsep}{2pt}
        \begin{tabular}{@{}ll|c@{}}
                            & {Method}         & \voc  \\ \shline
        \arch{a}            & Baseline                & 53.2      \\
        \arch{b}            & + Decoder         & 52.3      \\
        \grayrow \arch{c}   & + TCL        & \textbf{77.4}        \\
        \multicolumn{3}{c}{~} \\
        \multicolumn{3}{c}{~} \\
        \multicolumn{3}{c}{~} \\
        \end{tabular}
        \end{minipage}
    }
    \subfloat[
        {\small \textbf{\method losses}.}
        \label{table:ablations-gcl}
    ]{
        \begin{minipage}{0.53\columnwidth}
        \centering
        \small
        \renewcommand{\arraystretch}{1.05}
        \setlength{\tabcolsep}{2pt}
        \begin{tabular}{@{}ccccc|c@{}}
         & \head{$\text{\method}_v$} & \head{$\text{\method}_f$} & \head{$\mathcal{L}_\text{area}$} & \head{CL} & \voc   \\ \shline
        \arch{d}                   &          &         &         & \gcheck & 61.1                         \\
        \arch{e}                   & \gcheck  &         & \gcheck &         & 74.6                         \\
        \arch{f}                   &          & \gcheck & \gcheck &         & 76.0                         \\
        \grayrow \arch{c}          & \gcheck  & \gcheck & \gcheck &         & \textbf{77.4}                \\ 
        \arch{g}                   & \gcheck  & \gcheck & \gcheck & \gcheck & 75.6                \\
        \arch{h}                   & \gcheck  & \gcheck &         &         & 67.1                \\ 
        \end{tabular}
        \end{minipage}
    } 
    \\
    \vspace{0.25cm}
    \subfloat[
        {\small \lossmethod}
        \label{table:ablations-TCL}
    ]{
        \begin{minipage}{0.3\columnwidth}
        \centering
        \small
        \renewcommand{\arraystretch}{1.05}
        \setlength{\tabcolsep}{4pt}
        \scalebox{\scale}{
            \begin{tabular}{@{}c|c@{}} 
            $\lambda_\text{\method}$ & \voc   \\ \shline
            0.0 & -          \\
            0.01     & 76.8          \\
            \grayrow 0.1 & \textbf{77.4}          \\
            1.0      & 68.2         \\
            \end{tabular}
        }
        \end{minipage}
    }
    \subfloat[
        {\small \lossarea}
        \label{table:ablations-area}
    ]{
        \begin{minipage}{0.3\columnwidth}
        \centering
        \small
        \renewcommand{\arraystretch}{1.05}
        \setlength{\tabcolsep}{4pt}
        \scalebox{\scale}{
            \begin{tabular}{@{}c|c@{}} 
            $\lambda_\text{area}$ & \voc    \\ \shline
            0.0     & 67.1     \\
            0.04    & 69.5     \\
            \grayrow 0.4 & \textbf{77.4}          \\
            4.0     & 76.7     \\
            \end{tabular}
        }
        \end{minipage}
    }
    \subfloat[
        {\small \losstv}
        \label{table:ablations-tv}
    ]{
        \begin{minipage}{0.3\columnwidth}
        \centering
        \small
        \renewcommand{\arraystretch}{1.05}
        \setlength{\tabcolsep}{4pt}
        \scalebox{\scale}{
            \begin{tabular}{@{}c|c@{}} 
            $\lambda_\text{tv}$ & \voc    \\ \shline
            0.0   & 73.8      \\
            0.1   & 75.2     \\
            \grayrow 1.0 & \textbf{77.4}          \\
            10.0  & 70.7      \\
            \end{tabular}
        }
        \end{minipage}
    }
    \vspace{-0.05cm}
    \caption{\small \textbf{Ablation studies on \method losses and hyperparameters.} 
    Refinement techniques are not applied to reveal the effect of each loss function clearly.
    Default settings are marked in \colorbox{graycolor}{gray}.
    }
    \vspace{-0.32cm}
    \label{table:ablations}
\end{table}

\section{Conclusion}
We propose a novel framework for open-world semantic segmentation with only image-text pairs, addressing the alignment-level discrepancy between training (image-text) and testing (region-text) in existing methods. 
In the proposed framework, we incorporate the grounding process within contrastive learning, thus allowing explicitly learning alignment between text and text-grounded regions (\ie, segmentation mask).
We also present a unified evaluation protocol for a fair comparison of existing methods, where TCL achieves state-of-the-art zero-shot segmentation performance on all 8 benchmarks, remarkably surpassing previous methods.
We hope that this study encourages a new research direction of explicitly learning region-text alignment for open-world semantic segmentation.
%

\clearpage
{\small
\bibliographystyle{ieee_fullname}
\bibliography{egbib}
}

\clearpage
\appendix
\twocolumn[{%
    \renewcommand\twocolumn[1][]{#1}%
    \centering\noindent
    \scalebox{0.88}{
\begin{lstlisting}[
            language=Python,
            basicstyle=\ttfamily\small,
        ]
# x_t[B, L]       - minibatch of texts
# x_v[B, 3, H, W] - minibatch of images
# text_enc        - encode text (x_t) to text emb (t)
# image_enc       - encode image (x_v) to global image emb (v_g) and dense feature (v_d)
# grounding_dec   - decode dense feature (v_d) to pixel-level dense image emb (v_s)
# proj            - scalar projection and sigmoid layer
# gumbel_max      - Gumbel-Max function
# ld_area         - weight of area TCL loss

t = text_enc(x_t)  # t[B, C]
_, v_d = image_enc(x_v)  # v_d[B, L, C]
v_s = grounding_dec(v_d)  # v_s[B, C, H, W]
mask = proj(einsum("ichw,jc->ijhw", v_s, t))  # [B, B, H, W]

pos_mask = mask[arange(B), arange(B)].unsqueeze(1)  # [B, 1, H, W]
pos_mask_b = gumbel_max(pos_mask)  # binarized positive mask
v_g, _ = image_enc(pos_mask_b * x_v)  # v_g[B, C]
s = v_g @ t.T  # s[B, B]
tcl_v = info_nce(s)  # InfoNCE loss

w = mask / mask.sum((2, 3), keepdim=True)
grounded_emb = einsum("ichw,ijhw->ijc", v_s, w)
s = einsum("ijc,jc->ij", grounded_emb, t)  # s[B, B]
tcl_f = info_nce(s)  # InfoNCE loss

pos_area = pos_mask.mean()
neg_area = off_diag(mask).mean()  # average of off-diagonal
tcl_area = ld_area * ((pos_area - 0.4).abs() + neg_area)

tcl_loss = tcl_v + tcl_f + tcl_area
\end{lstlisting}
    }
    \vspace{-0.1cm}
    \captionof{figure}{\small \textbf{PyTorch-like pseudo code for the core implementation of \method.}}
    \label{fig:pseudocode}
    \vspace{0.7cm}
}]
\section{Pseudo-code}
\label{sec:pseudo-code}

For clarity, we present the pseudo-code for the core implementation of TCL in \cref{fig:pseudocode}. As described in \cref{sec:methods}, we first generate text-grounded masks via grounder and use them to compute text-grounded images and embeddings (L10-L17). 
Furthermore, this pseudo-code also demonstrates our efficiency-aware design of TCL.
We compute the $B\times B$ mask, where $B$ is the batch size, using a single \texttt{einsum} operation and scalar projection (L13).
When computing TCL$_v$ loss, we need to perform CLIP image encoder inference again for the grounded images (L17). To reduce computational complexity, we only use the positive mask, which is the diagonal of the quadratic mask, for linear inference with a size of $B$ instead of quadratic (L15).
When computing TCL$_f$ loss, we use the entire quadratic mask because a single \texttt{einsum} operation can efficiently compute the grounded embeddings without requiring additional encoder inference (L22).
A more detailed discussion on the efficiency is provided in the later section (\cref{sec:efficiency-analysis}).


\section{Architecture Details}
\label{sec:arch-details}

\begin{figure}[htpb]
    \centering
    \includegraphics[width=0.95\columnwidth]{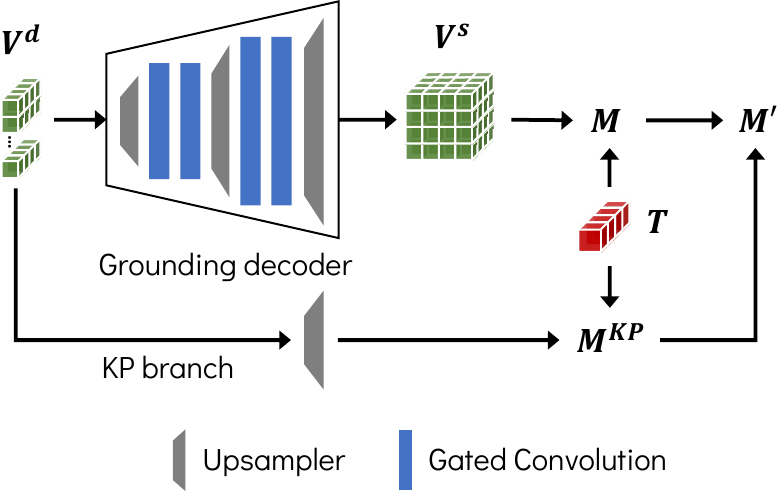}
    \caption{\small \textbf{Architecture of the grounding decoder.} The knowledge preservation (KP) branch serves to preserve pre-trained knowledge intact.}
    \label{fig:arch-detail}
\end{figure}

\begin{table*}[t]
\centering
\small
\renewcommand{\arraystretch}{1.0}
\setlength{\tabcolsep}{4pt}
    \begin{tabular}{l|c|ccc|ccccc|c}
    \toprule
         & & \multicolumn{3}{c|}{with background class} & \multicolumn{5}{c|}{without background class}     &       \\
        \textbf{Methods}  & {PAMR} & VOC  & {Context} & {Object}            & {VOC20} & {Context59} & Stuff & City & ADE  & Avg.  \\ 
        \midrule
        GroupViT        & & \underline{50.4} & 18.7 & \underline{27.5}           & \textbf{79.7}  & 23.4  & 15.3  & 11.1 & 9.2  & 29.4  \\
        MaskCLIP        & & 38.8 & \underline{23.6} & 20.6           & 74.9  & \underline{26.4}  & \underline{16.4}  & 12.6 & 9.8  & 27.9  \\
        ReCo            & & 25.1 & 19.9 & 15.7           & 57.7  & 22.3  & 14.8  & \underline{21.1} & \underline{11.2} & 23.5  \\
        TCL (Ours)             & & \textbf{51.2} & \textbf{24.3} & \textbf{30.4}           & \underline{77.5}  & \textbf{30.3}  & \textbf{19.6}  & \textbf{23.1} & \textbf{14.9} & \textbf{33.9}  \\
        %
        \midrule
        GroupViT & \ding{52} & \underline{51.1} & 19.0 & \underline{27.9}           & \underline{81.5}  & 23.8  & 15.4  & 11.6 & 9.4  & 30.0  \\ 
        MaskCLIP & \ding{52} & 37.2 & \underline{22.6} & 18.9           & 72.1  & \underline{25.3}  & 15.1  & 11.2 & 9.0  & 26.4  \\ 
        ReCo     & \ding{52} & 27.2 & 21.9 & 17.3           & 62.4  & 24.7  & 16.3  & \underline{22.8} & \underline{12.4} & 25.6  \\ 
        TCL (Ours)     & \ding{52} & \textbf{55.0} & \textbf{30.4} & \textbf{31.6}           & \textbf{83.2}  & \textbf{33.9}  & \textbf{22.4}  & \textbf{24.0} & \textbf{17.1} & \textbf{37.2}  \\
        \bottomrule
    \end{tabular}%
\vspace{-0.1cm}
\caption{\small 
\textbf{Standardizing PAMR condition.}
To reveal the effect of our refinement method (PAMR), we compare the methods with the same PAMR condition. TCL achieves the best in both with and without PAMR settings.
}
\label{table:pamr}
\vspace{-0.1cm}
\end{table*}

Our core design principle is to preserve and exploit the diverse knowledge of pre-trained CLIP \cite{radford2021clip}. 
Therefore, we freeze the pre-trained CLIP encoders\footnote{After $30,000$ iterations, we unfreeze only the last block of the image encoder for richer model capability.} and train the grounding decoder for the adaptation from the image-text alignment to the region-text alignment.
We also considered the other techniques to preserve knowledge \cite{cha2022miro}, but a simple freezing strategy worked the best.
We follow the simple modification of MaskCLIP \cite{zhou2022maskclip} to the CLIP image encoder. They modify the last attention of the CLIP image encoder to acquire the dense features representing local semantics. These dense image features $\mathbf V^d$ are fed to the grounding decoder. As shown in \cref{fig:arch-detail},
the grounding decoder consists of four gated convolution blocks, where the output of a convolution is gated by a learned gating parameter and added to the skip connection. Concretely, the process of gated convolution can be written as: 
\begin{equation}
    \mathbf x' = \mathbf x + \text{tanh}(g) \cdot \text{Conv}(\mathbf x),
\end{equation}
where $\mathbf x$ is input feature and $g$ is a learned gating parameter. The upsamplers increase the feature map resolution for the high-resolution segmentation capability. The first two upsamplers use the nearest neighbor interpolation and the last upsampler adapts the resulting embedding into the pixel-level embedding by the bilinear interpolation.
In addition, as shown in \cref{fig:arch-detail}, we employ two branches strategy: the main grounding decoder and knowledge preservation (KP) branches.
In this KP branch, the CLIP dense features $\mathbf V^d$ are reshaped spatially and upsampled to pixel-level resolution by bilinear interpolation, and then we compute the text-grounded mask $\mathbf M^\text{KP}$ by \cref{eq:pred-mask}.
There are no learnable parameters in this branch and the output masks are just mixed with the output from the grounding decoder branch as follows:
\begin{equation}
    \mathbf M' = (1 - w_{kp}) \cdot \mathbf M + w_{kp} \cdot \mathbf M^\text{KP},
\end{equation}
where $\mathbf M^\text{KP}$ is the generated masks from knowledge preservation branch, $\mathbf M'$ is the final output mask, and $w_{kp}$ is a mixing hyperparameter. 
We use the $w_{kp}$ of 0.3. To fully leverage the massive pre-trained knowledge of CLIP \cite{radford2021clip}, this branch is only used in the inference stage. It also can be regarded as a cost-free ensemble.


\section{Fair Comparison}
\label{sec:fair_comparison}
In this paper, we present a unified evaluation protocol to facilitate a fair and rigorous comparison. However, the condition of fair evaluation protocol can be controversial. Thus, in this section, we provide additional comparisons to enhance fairness, paving the way for future research on fair comparison in open-world segmentation.

\paragraph{Fair comparison with refinement methods.}
In our unified evaluation protocol, we do not unify refinement methods, as we believe that each method's approach to refining the model output is a design choice. However, some may argue that a fair comparison protocol should unify the refinement methods as well.
To address this concern, we also provide the comparison with and without PAMR \cite{araslanov2020pamr}, which is our refinement method used in TCL.
As shown in \cref{table:pamr}, even using the same refinement method across all methods, TCL still achieves state-of-the-art performance with a significant margin in both settings, demonstrating the effectiveness of its underlying approach.
It is also worth noting that the performance gains resulting from PAMR are specific to each method. 
For example, TCL and ReCo demonstrate significant performance improvements of $+3.3$ and $+2.1$ mIoU, respectively, while GroupViT only shows a marginal gain of $+0.6$ mIoU, and MaskCLIP actually leads to a decrease in performance of $-1.5$ mIoU.
We believe that this is because each comparison method was designed without considering refinement by PAMR. Thus, because the effectiveness of the refinement method is closely tied to the main method, we have not unified the use of refinement methods in our proposed evaluation protocol.


\begin{table}[t]
\centering
\small
\renewcommand{\arraystretch}{1.1}
\setlength{\tabcolsep}{3pt}
\resizebox{1.0\columnwidth}{!}{%
    \begin{tabular}{@{}l|l|cc|c@{}}
    \toprule
    \textbf{Methods}          & {Datasets}  & {VOC20} & {Context59} & Avg.  \\ \midrule
    \multirow{4}{*}{GroupViT} & CC15M + RedCaps12M & \underline{79.7}           & 23.4  & \underline{51.6}                \\
                              & CC15M + YFCC14M    & 74.1           & 20.8  & 47.5               \\
                              & CC15M + Coyo100M   & 73.3           & \underline{25.0}  & 49.2                \\
                              & CC15M + Coyo700M   & 75.5           & 24.2  & 49.9                \\ \midrule
    TCL (Ours)                 & CC15M + WIT400M    & \textbf{83.2}           & \textbf{33.9}  & \textbf{58.6}            \\ \bottomrule
    \end{tabular}
}
\vspace{-0.1cm}
\caption{\small 
    \textbf{Scale-up GroupViT}
    does not directly help the segmentation capability. WIT400M indicates the dataset of CLIP pre-training.
}
\label{table:scaleup-groupvit}
\vspace{-0.1cm}
\end{table}
\paragraph{Fair comparison in dataset scale.}
Except for GroupViT, all comparison methods (TCL, ReCo, and MaskCLIP) utilize CLIP pre-trained models. 
GroupViT proposes a new encoder architecture and is therefore unable to leverage CLIP pre-trained models directly, which is one of its limitations. 
Hence, a fair comparison would be to evaluate GroupViT as it is. 
Nevertheless, it could be argued that comparing GroupViT and other methods at the same training dataset scale would be a fair comparison. To address this concern, we provide additional scale-up experiments. As mentioned, GroupViT is unable to leverage CLIP pre-trained models directly. 
Thus, for a fair comparison, we train GroupViT using the publicly available Coyo700M dataset of large-scale image-text pairs \cite{kakaobrain2022coyo-700m}, which is larger than the non-public CLIP training dataset (WIT400M).
However, we observe that a simple scale-up of the training dataset does not guarantee improvement in performance. 
As shown in \cref{table:scaleup-groupvit}, the correlation between dataset size and performance is not clear. The impact of larger datasets varies between the datasets (\eg, ``CC15M+RedCaps12M'' performs best on VOC20, but ``CC15M+Coyo100M'' performs best on Context59), and TCL outperforms all variants of GroupViT despite the fair dataset scale.


\begin{table}[t]
\centering
\small
\renewcommand{\arraystretch}{1.2}
\setlength{\tabcolsep}{3pt}
\scalebox{0.96}{
    \begin{tabular}{@{}l|ccccc@{}} 
    \toprule
                                          & \method & $\text{TCL}^\dagger$ & GroupViT & MaskCLIP & ReCo  \\ \midrule
    Speed (s)     & 0.08 & 0.07 & 0.05    & 0.04     & 28.10  \\
    FPS (it/s)    & 12.93 & 15.11 & 20.94    & 26.09     & 0.04   \\
    Mask ratio    & $1/4^2$ & $1/4^2$ & $1/16^2$ & $1/16^2$ & $1/16^2$ \\
    \bottomrule
    \end{tabular}
}
\caption{\small \textbf{Inference speeds and FPS.} 
Mask ratio indicates the resolution ratio of the segmentation masks compared to the input image size. For example, TCL generates $112\times 112$ masks for $448\times 448$ input image, while GroupViT generates $28\times 28$ masks.
$\text{TCL}^\dagger$ denotes the TCL model without PAMR.
}
\label{table:throughput}
\vspace{-0.3cm}
\end{table}
\section{Efficiency Analysis}
\label{sec:efficiency-analysis}
While efficiency is not the primary objective of this study, it is also considered one of our core design principles, especially regarding inference efficiency for practical applications. This section provides an analysis of the inference throughput and our design choices for efficiency.

\paragraph{Inference throughput.}
\label{sec:infer-throughput}
We benchmark the inference speeds and FPS using $448 \times 448$ images and $21$ target classes on a single NVIDIA V100 GPU. Our benchmark setting follows an open-world scenario that addresses an arbitrary class, meaning that the text embeddings are computed for every inference. 
%
As shown in \cref{table:throughput}, TCL has slightly lower FPS compared to GroupViT or MaskCLIP, due to the relatively high resolution of segmentation masks. This trade-off between mask resolution and FPS can be controlled by the design of the grounding decoder, but we do not investigate this further, as it is outside the scope of this study.
ReCo \cite{shin2022reco} shows very low 
FPS due to its retrieval process, which involves retrieving similar images from the ImageNet dataset \cite{russakovsky2015imagenet} and co-segmenting them. 
Note that there are many methods to reduce the computational cost of TCL, \eg, utilizing depthwise convolution instead of standard convolution \cite{chollet2017xception}, but it is not the focus of this study and is left for future work.

\paragraph{Efficiency-aware designs in TCL.}
In contrastive learning, the image-text alignment requires $B \times B$ matching, where $B$ denotes either the batch size in training or the number of target classes in testing.
This quadratic operation is a major bottleneck when scaling up the batch size. To address this, CLIP \cite{radford2021clip} uses a late fusion architecture to minimize the $B \times B$ operation. In this architecture, the encoders are decoupled, and the encoded embeddings are combined at the end of the network, making the model efficient in both training and inference.
We also employ the late fusion architecture following CLIP for efficiency.
In addition, since TCL$_v$ requires additional encoding of text-grounded images ($\tilde{\mathbf x}^V$), we use $\tilde{\mathbf x}^V$ from only positive pairs (=$B$ encodings), and bypass $B\times B$ encodings by introducing TCL$_f$ to compute $\mathbf v^f$ without any additional encoding.
We further improve efficiency by freezing encoders and reusing text embeddings.
As a result, our entire training time is about 12 hours, and our inference throughput is comparable to MaskCLIP or GroupViT.
Nevertheless, it is also worth noting that scaling up TCL is more challenging than CLIP since the $B \times B$ mask is a spatial tensor ($B \times B \times H \times W$), unlike CLIP. Empirically, we found that algorithmic improvements on the scale should be considered to scale up to $B > 8,192$.
%

\section{Additional Details and Experiments}
\label{sec:additional-comparison-and-details}

\begin{figure}[t]
    \centering
    \includegraphics[width=\linewidth]{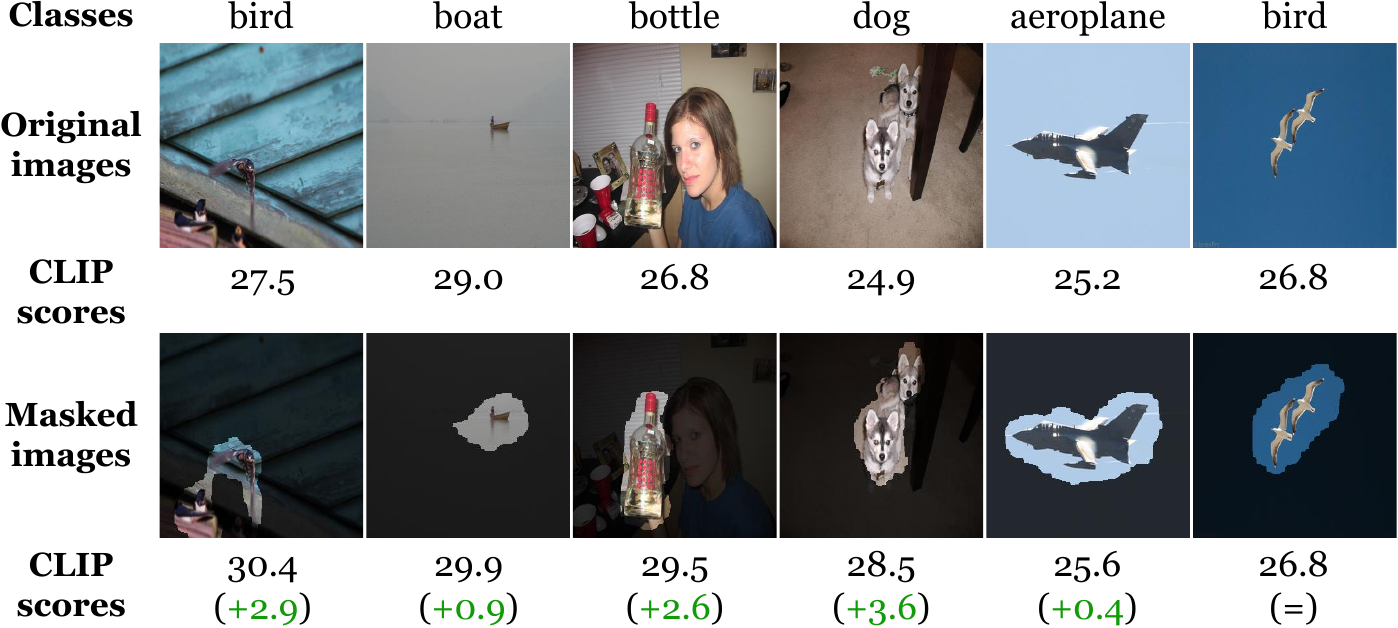}
    \caption{
        \small \textbf{Qualitative examples on the robustness of CLIP for masked images.}
        The first and second rows show natural (non-masked) and masked images, respectively. The number below each image indicates CLIP score between the image and the corresponding class.
    }
    \vspace{-0.1cm}
    \label{fig:masked-images-with-clip}
\end{figure}
\paragraph{CLIP with masked images.}
TCL implicitly assumes that CLIP can address masked images robustly. While CLIP is widely known for its strong robustness \cite{radford2021clip}, 
it is not yet clear how well CLIP can handle masked images.
As such, we investigate CLIP scores between masked images and positive texts in \cref{fig:masked-images-with-clip}. The results indicate that CLIP can address masked images. Interestingly, the scores tend to be improved, especially for the images with complex context (columns 1, 3, and 4 of \cref{fig:masked-images-with-clip}). The masks remove noisy context and help to focus on the target object, leading to improved scores.

\paragraph{Details on the comparison methods.}
In the quantitative evaluation, we include the variants of the comparison baselines for an extensive comparison; the GroupViT variants from the YFCC and RedCaps checkpoints \cite{xu2022groupvit} and the MaskCLIP variants by the refinement process (key smoothing and prompt denoising). For the backbone of MaskCLIP, we use ViT-B/16 since its reported performance is better than ResNet-50 \cite{zhou2022maskclip}. For the qualitative comparison, we choose the quantitatively best variant for each method.

\begin{table}[t]
\centering
\small
\renewcommand{\arraystretch}{1.1}
\setlength{\tabcolsep}{3pt}
\begin{tabular}{l|ccc|c} 
\toprule
        \textbf{Methods}    & VOC20         & Context59     & Stuff    & Avg.           \\ \midrule
ViL-Seg             & 34.4          & 16.3          & 16.4          & 22.4           \\
ReCo               & 57.9          & 32.0          & 18.4          & 36.1           \\
GroupViT {\scriptsize (RedCaps)}  & 79.0          & 49.2          & 16.1          & 48.1           \\
MaskCLIP  & 73.0          & 56.5          & 21.9          & 50.5           \\ \midrule
\method (Ours) & \textbf{84.5} & \textbf{62.0} & \textbf{27.6} & \textbf{58.0}  \\
\bottomrule
\end{tabular}
\caption{\small \textbf{Zero-shot segmentation performance for partial classes.} The numbers of ViL-Seg are adopted from the original paper \cite{liu2022vilseg}. We use the proposed unified evaluation protocol for the other results.
}
\label{table:zeroshot_partial}
\vspace{-0.2cm}
\end{table}

\paragraph{Comparisons with zero-shot semantic segmentation methods.}
We provide an extensive and unified comparison in \cref{table:main}, but the comparison does not include non-open-sourced methods. To the best of our knowledge, ViL-Seg \cite{liu2022vilseg} is the only non-open-sourced method for open-world semantic segmentation. We compare the zero-shot segmentation performance following the evaluation protocol of ViL-Seg. 
In this evaluation protocol for zero-shot semantic segmentation, only partial classes are used: 5 classes (potted plant, sheep, sofa, train, tv-monitor) for PASCAL VOC20, 4 classes (cow, motorbike, sofa, cat) for PASCAL Context59, and 15 classes (frisbee, skateboard, cardboard, carrot, scissors, suitcase, giraffe, cow, road, wall concrete, tree, grass, river, clouds, playingfield) for the COCO-Stuff dataset. Therefore, we additionally provide the comparison results under the partial classes protocol. As shown in \cref{table:zeroshot_partial}, \method achieves state-of-the-art performance with a large margin in every dataset again.


\section{Case Study on Failures}
\label{sec:failure-case-study}

\begin{figure*}[t]
    \centering
    \begin{subfigure}{0.44\linewidth}
        \centering
        \includegraphics[width=\linewidth]{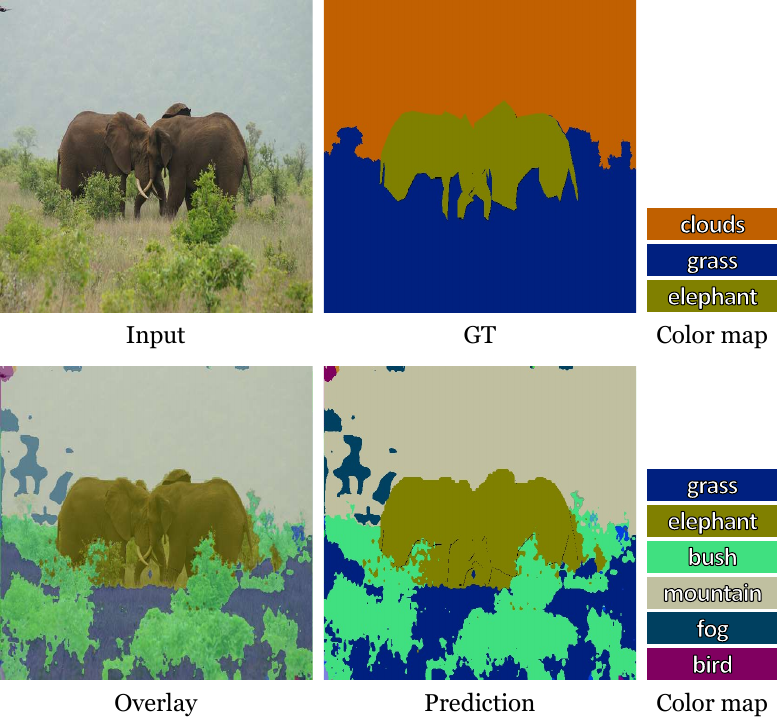}
        \caption{\small {Case 1}}
        \label{fig:failure-case-1}
    \end{subfigure}
    \hspace{0.3cm}
    \begin{subfigure}{0.518\linewidth}
        \centering
        \includegraphics[width=\linewidth]{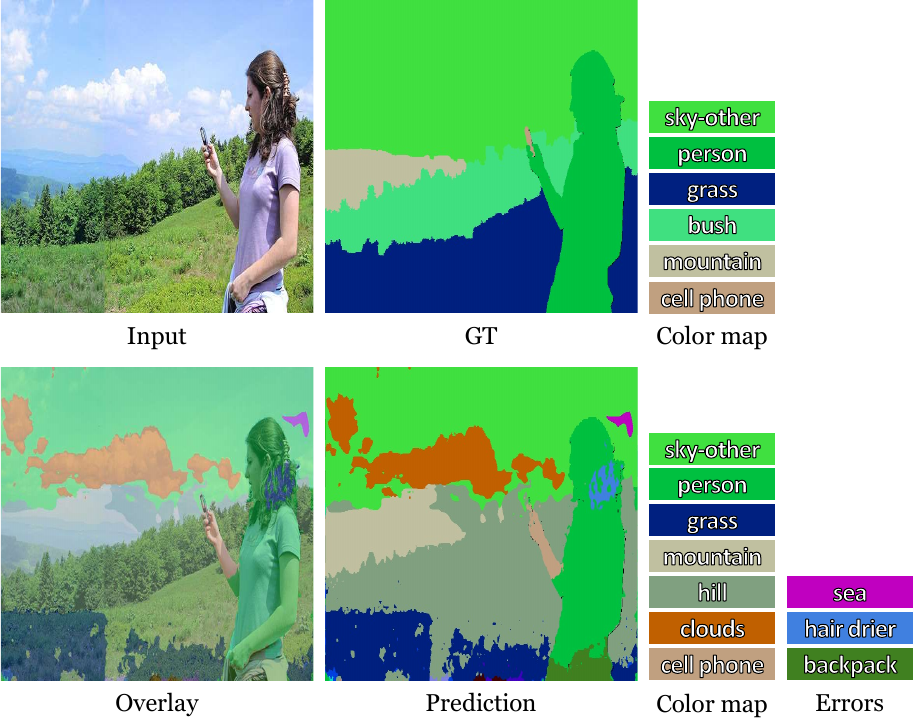}
        \caption{\small {Case 2}}
        \label{fig:failure-case-2}
    \end{subfigure}
    \vspace{-0.1cm}
    \caption{\small \textbf{Ambiguous error cases.}
    The color map of a negligible region is omitted. Although different from the ground truth (GT), there are many cases that can be considered correct.
    The segments in the color map are semantically correct despite of difference with GT, while the errors indicate clearly incorrect predictions.
    }
    \label{fig:failure-cases}
    \vspace{-0.1cm}
\end{figure*}

\paragraph{Errors of \method.} We investigate the failure cases of \method via various qualitative examples. First, \method undergoes difficulty in capturing segment boundaries accurately. For example, in \cref{fig:failure-case-2}, the predicted segment of the ``mountain'' class includes part of the sky region, and the ``cell phone'' segment contains the right hand and arm regions. This is a fundamental challenge of unsupervised open-world segmentation; the absence of dense annotation makes precisely capturing a segment boundary extremely difficult. 
Although the proposed method remarkably improves the segmentation performance compared with previous methods, this case study reveals that there are still many areas to be improved. Furthermore, despite the help of the smooth prior loss, the predictions still tend to be noisy, \eg, ``sea'' or ``hair drier'' in \cref{fig:failure-case-2}.

\paragraph{Ambiguity in benchmarks.}
On the other hand, we also find crucial issues in the current benchmark datasets: ambiguities in the class label set and scene semantics. In particular, there are lots of class labels with similar semantics, especially in the datasets with a large vocabulary, \eg, COCO-Stuff (171 classes) \cite{caesar2018cocostuff} or ADE20K (150 classes) \cite{zhou2019ade20k}. Mostly the labels have different semantics in detail, but the distinction between the labels can be ambiguous depending on how the image captures the scene. 
For example, it is hard to distinguish ``clouds'' and ``fog'' in \cref{fig:failure-case-1} and ``hill'' and ``mountain'' in \cref{fig:failure-case-2}. 
Also, there are labels with superset-subset relations. For instance, the COCO-Stuff dataset has ``broccoli'', ``vegetable'', and ``food-other'' classes. In the supervised setting, a model can address this issue by training only if there is labeling consistency between images. However, in the open-world scenario, such superset-subset relations cause significant ambiguity.
Furthermore, a more frequent ambiguity raises when a segment has multiple semantics. 
More proper descriptions of the ``clouds'' and ``grass'' segments in \cref{fig:failure-case-1} are ``foggy or cloudy mountain'' and ``bushes on the grass'', respectively. However, ground truth (GT) labels represent only part of the entire semantics. As with the superset-subset relation case, benchmarks for the open-world scenario require additional consideration to address such ambiguities.
In this study, we propose a unified evaluation protocol to compare the existing methods fairly, but it only unifies the evaluation protocol and simply employs existing benchmark datasets. This analysis suggests the need for further advanced benchmarks dedicated to open-world scenarios in the future.


\begin{figure}[t]
    \centering
    \includegraphics[width=\linewidth]{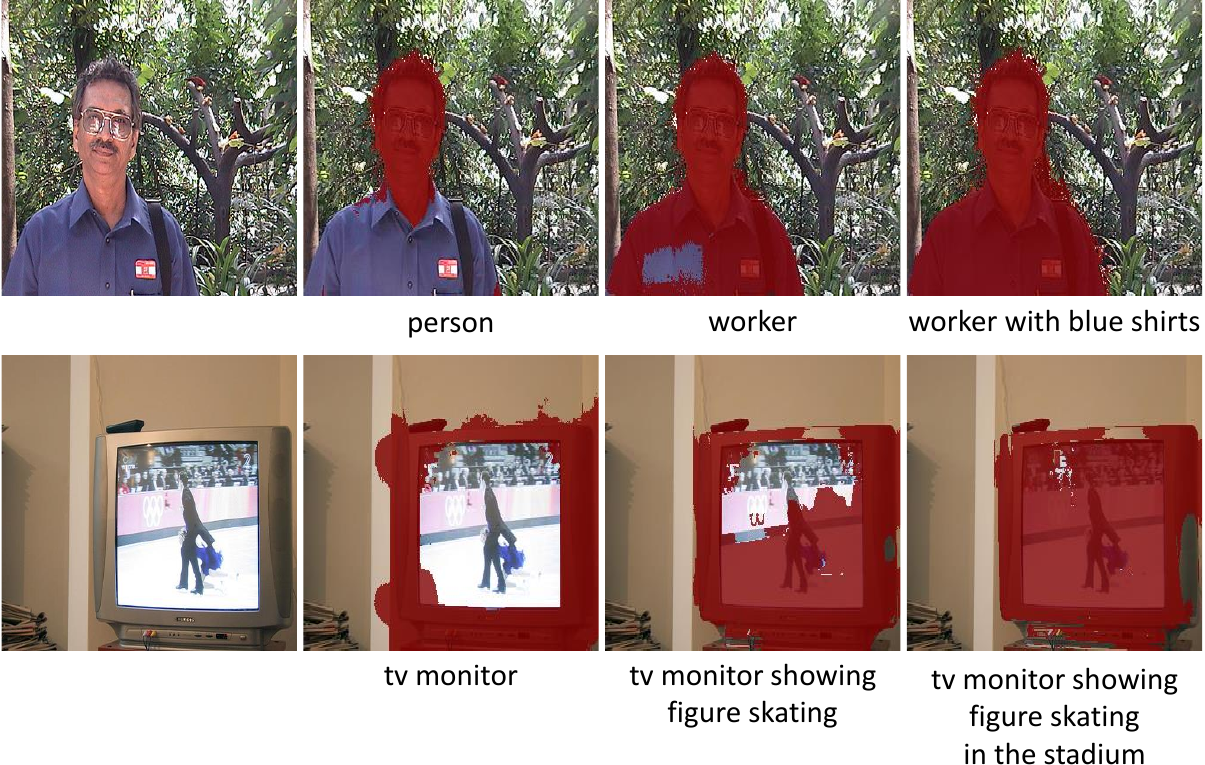}
    \caption{\small \textbf{\method behaviors depending on the specificity of the prompts.}
    The red-colored region indicates the segmentation results of the given text prompts. The more specific the prompt, the better the segmentation result.
    }
    \vspace{-0.1cm}
    \label{fig:behavior-analysis}
\end{figure}
\begin{figure*}[t]
    \centering
    \includegraphics[width=0.9\linewidth]{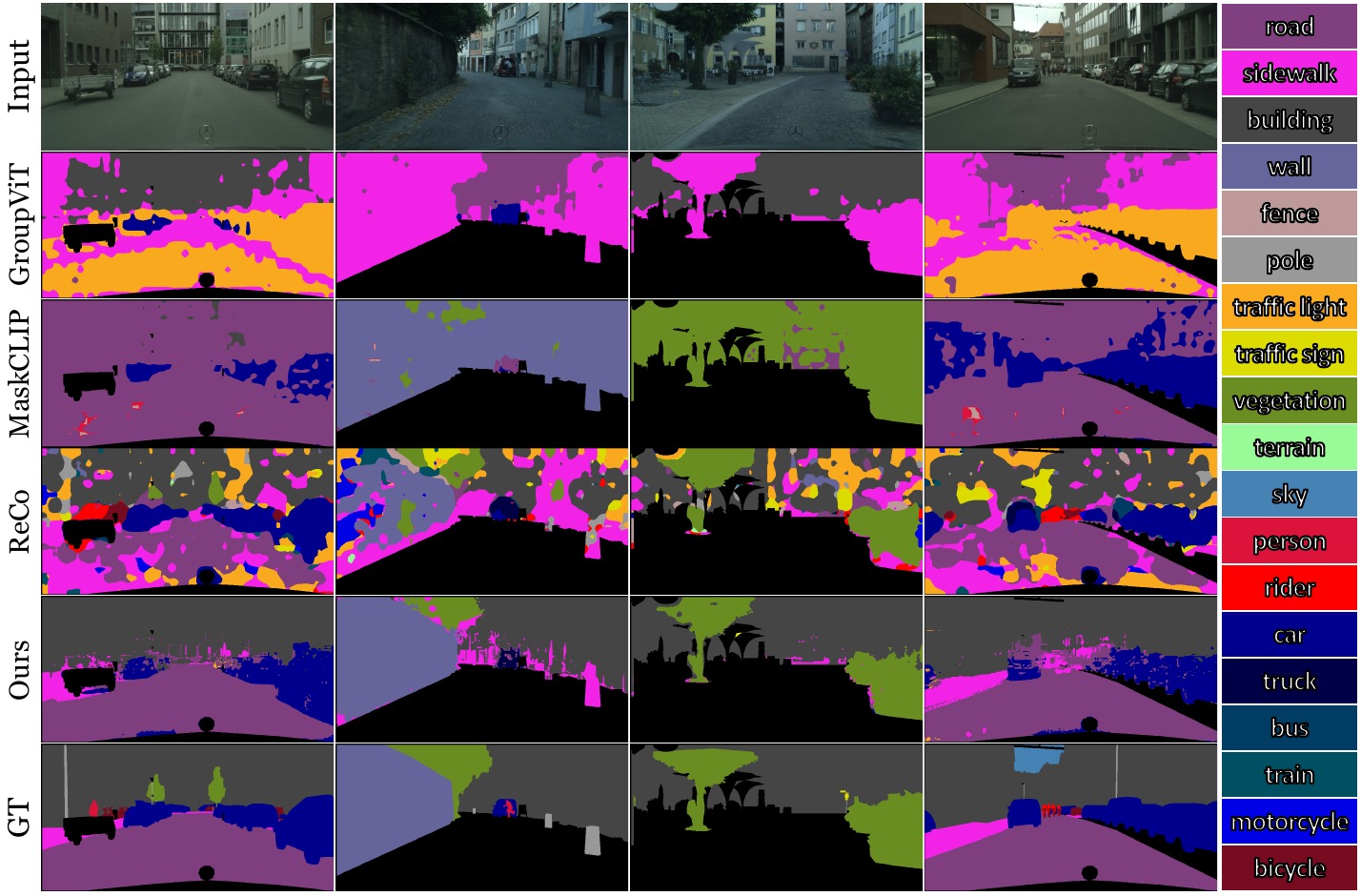}
    \vspace{-0.15cm}
    \caption{\small \textbf{Additional qualitative examples on Cityscapes.}
    }
    \label{fig:qualitative-city}
\end{figure*}

\section{Analysis on Model Behavior}
\label{sec:behavior-analysis}

In this section, we investigate how the learned \method model generates different segmentation masks depending on the input text prompts. As shown in \cref{fig:behavior-analysis}, the model tends to capture the intended region better when the input text prompt is more specific. Although this characteristic can cause performance degradation when evaluating the model on a fixed benchmark, it also improves the controllability of the model. 
We can exploit this controllability to maximize the benchmark performance, \eg, class name expansion. However, we do not employ these dataset-dependent tricks to prevent the overestimation of the model performance, as described in \cref{subsec:experiment-settings}.

\section{Additional Qualitative Results}
\label{sec:additional-qualitative}

\subsection{Qualitative Examples on Complicated Scene}

\begin{figure*}[ht]
    \centering
    \includegraphics[width=0.9\linewidth]{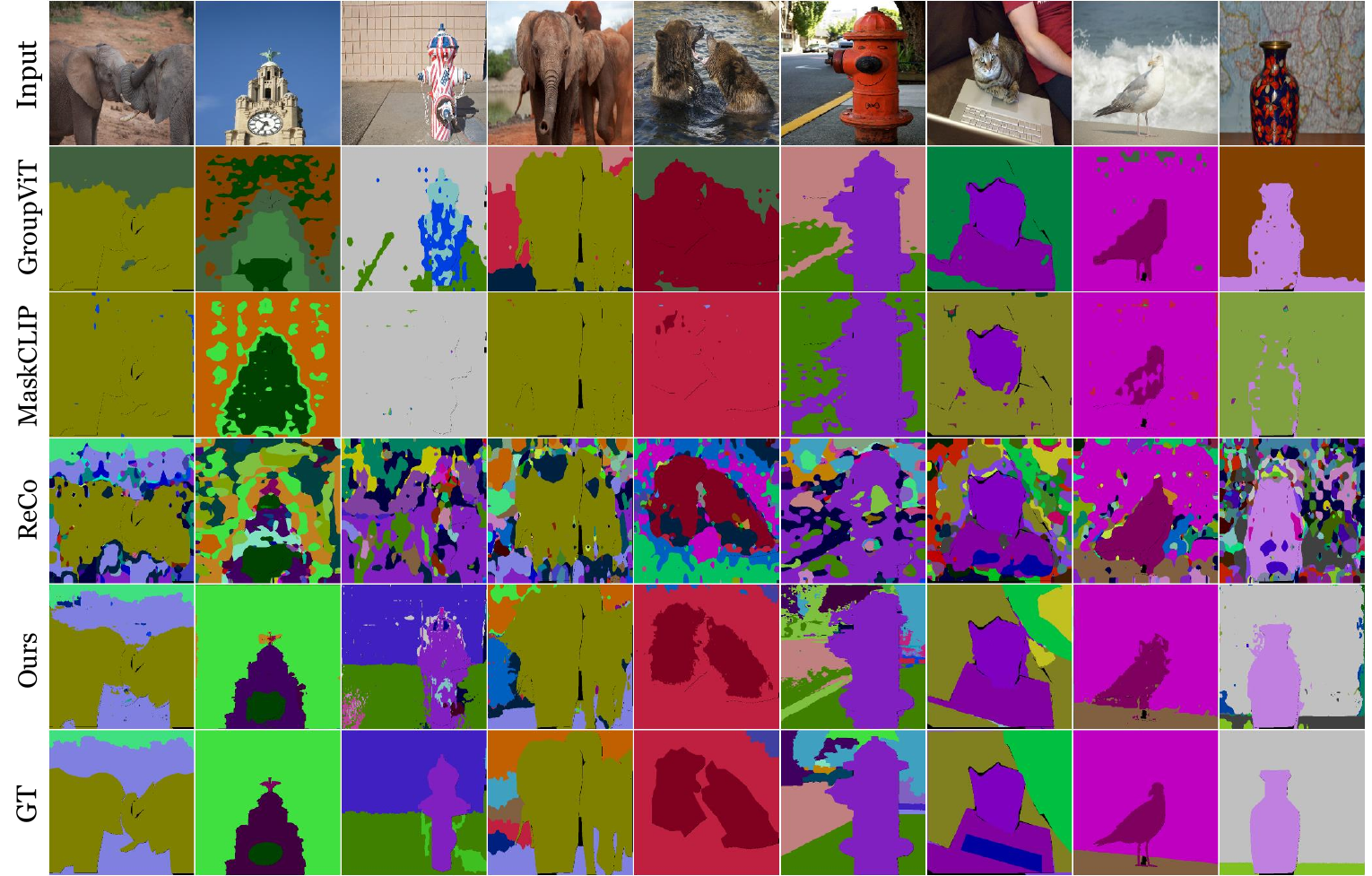}
    \vspace{-0.15cm}
    \caption{\small \textbf{Additional qualitative examples on COCO-Stuff.} 
    Color map is omitted since COCO-Stuff has 171 classes.
    }
    \label{fig:qualitative-stuff}
\end{figure*}

Qualitative comparison on PASCAL VOC in \cref{subsec:qualitative-results} visualizes the performance difference between comparison methods. 
However, the VOC dataset tends to be object-oriented and its images are generally composed of one or two segments.
In this section, we qualitatively compare open-world segmentation methods in complicated scenes. \cref{fig:qualitative-city,fig:qualitative-stuff} show examples including at least 3 segments from the Cityscapes and COCO-Stuff datasets. 
As shown in the figures, GroupViT \cite{xu2022groupvit} and MaskCLIP \cite{zhou2022maskclip} tend to generate a small number of segments. It makes the results less noisy but causes a large error. For example, in \cref{fig:qualitative-city}, MaskCLIP fails to segment ``building'' regions and GroupViT misidentifies ``road'' as ``traffic light''. In contrast, ReCo \cite{shin2022reco} suffers from noisy prediction. Our \method also generates partially noisy results, but it is relatively cleaner and better than the comparison baselines.
It is also worth noting that the image resolution of the unified evaluation protocol is relatively smaller than the widely used protocols for Cityscapes.
We resize a shorter side of an image to $448$ with keeping the aspect ratio, resulting in $448 \times 896$. In contrast, $1024 \times 2048$ is the widely used resolution for the Cityscapes dataset \cite{cheng2021maskformer,xie2021segformer}. 
Increasing the resolution can help recognize small objects, \eg, the persons in \cref{fig:qualitative-city}.

\subsection{Additional Qualitative Examples in the Wild}

\begin{figure*}[ht]
    \centering
    \includegraphics[width=0.9\linewidth]{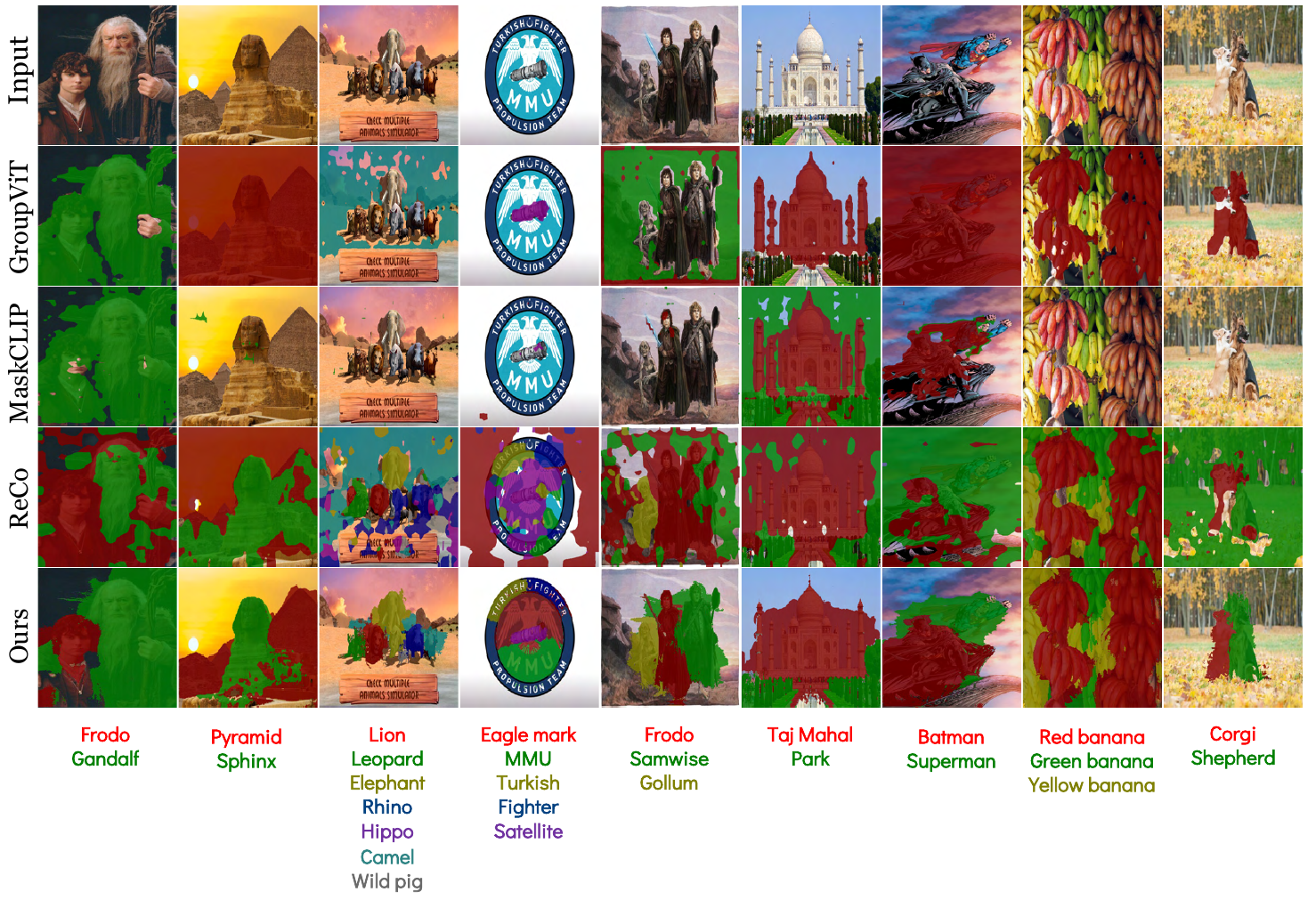}
    \caption{\small \textbf{Additional qualitative examples in the wild.} 
    }
    \label{fig:qualitative-web2}
\end{figure*}

\cref{fig:qualitative-web2} shows additional qualitative examples from web images in the wild. In this experiment, we investigate the discrimination capability of the model in various aspects: proper nouns (Frodo, Gandalf, Pyramid, Sphinx, Samwise, Gollum, Taj Mahal, Batman, Superman), colors with the same object (red, green, yellow bananas), letters (MMU, Turkish, Fighter), and subclasses (Corgi, Shepherd). The results show that our model can recognize and segment various concepts in the wild. For the baseline models, the results show similar tendencies with the in-the-wild examples in \cref{subsec:qualitative-results}. Contrary to the quantitative evaluation in fixed benchmarks, ReCo \cite{shin2022reco} generates a relatively plausible segmentation map compared with GroupViT \cite{xu2022groupvit} and MaskCLIP \cite{zhou2022maskclip}.


\end{document}